\documentclass[conference]{IEEEtran}
\IEEEoverridecommandlockouts
\usepackage{booktabs}
\usepackage{hyperref}
\usepackage[]{algorithm} 
\usepackage{algpseudocode}

\usepackage{cite}
\usepackage{amsmath,amssymb,amsfonts}

\usepackage{graphicx}
\usepackage{textcomp}
\usepackage{xcolor}
\usepackage{multirow}
\def\BibTeX{{\rm B\kern-.05em{\sc i\kern-.025em b}\kern-.08em
    T\kern-.1667em\lower.7ex\hbox{E}\kern-.125emX}}
\usepackage{fancyhdr}
\begin{document}

\title{Toward individual-level calibration in affect recognition with perceptual adjustment queries\\
}

\author{\IEEEauthorblockN{1\textsuperscript{st} Xuanzhou Chen}
\IEEEauthorblockA{\textit{Electrical and Computer Engineering} \\
\textit{Georgia Institute of Technology}\\
Atlanta, United States \\
xchen920@gatech.edu}
\and
\IEEEauthorblockN{2\textsuperscript{nd} Sankaraleengam Alagapan}
\IEEEauthorblockA{\textit{Electrical and Computer Engineering} \\
\textit{Georgia Institute of Technology}\\
Atlanta, United States \\
sankar.alagapan@gatech.edu}
\and
\IEEEauthorblockN{3\textsuperscript{rd} Ashwin Pananjady}
\IEEEauthorblockA{\textit{Industrial and Systems Engineering} \\
\textit{Electrical and Computer Engineering} \\
\textit{Georgia Institute of Technology}\\
Atlanta, United States \\
ashwinpm@gatech.edu}
}

\maketitle
\thispagestyle{fancy}

\begin{abstract}
    Behavioral tasks measuring facial affect perception assume that identical stimuli impose equivalent perceptual difficulty across participants. However, this assumption is systematically violated by individual differences in perceptual sensitivity. Using an affective perception task as our testbed, we propose a framework to normalize for perceptual difficulty that directly estimates each participant's Just Noticeable Difference (JND) along the facial affect spectrum via cognitively lightweight perceptual adjustment queries (PAQs). We use these PAQ-inferred JNDs to re-express stimulus distances, constructing  difficulty-equated tasks in perceptual space. We validate the framework in a Two-Alternative Forced-Choice (2AFC) task using two complementary behavioral measures: binary metacognitive difficulty judgments and response time variance decomposition. We find that PAQ calibration significantly equalizes perceived task difficulty at an individual level when compared to both the non-calibrated baseline and population-level Weibull calibration, while also reducing mean response time and between-subject variance in response time. These results establish PAQ as a principled and practical instrument for individualized perceptual calibration in facial affect recognition. Code: \url{https://anonymous.4open.science/r/PAQ_affect_normalization-BC67}.
\end{abstract}

\begin{IEEEkeywords}
Perceptual calibration, Human query mechanism, Affect recognition, Vision-language model, Facial generation.
\end{IEEEkeywords}

\section{Introduction}
Individuals perceive the world differently, and have non-uniform perception on any affective scale~\cite{binetti2022genetic, murray2024expression}. When a system assumes a uniform mapping between its internal representation and 
humans' perceptual scale, it incurs a \textit{perceptual miscalibration}: a 
systematic mismatch between what the system represents and what the individual actually perceives. 
Such perceptual miscalibration can systematically distort downstream behavioral or clinical analyses~\cite{hedge2018reliability}. It is especially harmful in tasks where the eventual scientific conclusion is based on a hypothesis about users' perceptual judgment along some affect spectrum. For example, applications in depression diagnosis~\cite{fan2024brain} postulate that depressed patients exhibit affective bias when presented with facial stimuli expressing different emotions: For a depressed patient, it is more cognitively demanding to distinguish between ``happy" faces than ``sad" ones, reflecting in that individual's response time to various questions. However, the two happy faces in the data set may be intrinsically more difficult to distinguish (because they are similar to each other) than the two sad faces, and calibrating for such differences in task difficulty is essential to answer the scientific question at hand.
While the difficulty of tasks can be normalized at a population-level using preliminary data collection, this approach fails to accommodate individual-level differences~\cite{gurler2015link,nath2012neural,takeuchi2017individual} and the analysis is not generalizable to individuals who deviate from the population.
Perceptual miscalibration arises in several applications of affective computing and human-machine interaction. Two representative cases are affect recognition and human-bot interaction:

\paragraph{Calibrating for individual differences in affect perception}
In affect recognition tasks, participants exhibit distinct 
perceptual scales even when presented with physically identical stimuli. For some participants, a 
small facial morphing increment produces a salient, noticeable emotional shift, while for 
others, the same increment remains nearly imperceptible. These variations reflect 
differences in perceptual sensitivity and internal normalization: each participant 
operates on a personalized scale mapping physical stimulus intensity 
to perceived emotional strength~\cite{murray2024expression, gao2014new}. There are two causes: i) A ``neutral" baseline is not universally defined~\cite{lee2008neutral, rohrbeck2023trait, suess2015perceiving}; ii) Individual response scalings are not calibrated~\cite{binetti2022genetic, murray2024expression}. Normalization in perceptual scale is hence essential to make inferences about how individuals perceive an affect.

\paragraph{Developing `personalized' human-bot gaming interaction} In cooperative strategy games such as Dota 2~\cite{dota2bots}, difficulty is partitioned into three discrete tiers (Easy, Medium, Hard). Nevertheless, players differ substantially in their ability to perceive and react in game events. For example, the same bot behavior that is trivially predictable for an experienced player may be cognitively overwhelming for a novice~\cite{zohaib2018dynamic}. The paper \cite{tran2025towards} shows that static difficulty systems fail to meet the needs of different types of players, since the perception of difficulty of a player depends on individual strengths, weaknesses, and personality traits that objective performance metrics alone cannot capture. Perceptual miscalibration occurs since the fixed difficulty levels of the gaming system are not tailored to each player's perceived challenge level.

\subsection*{Setup and contributions}
In both cases above,  miscalibration occurs because the system lacks access to the individual's perceptual model and has no mechanism for inferring it. 

\begin{figure}
    \centering
    \includegraphics[width=\linewidth]{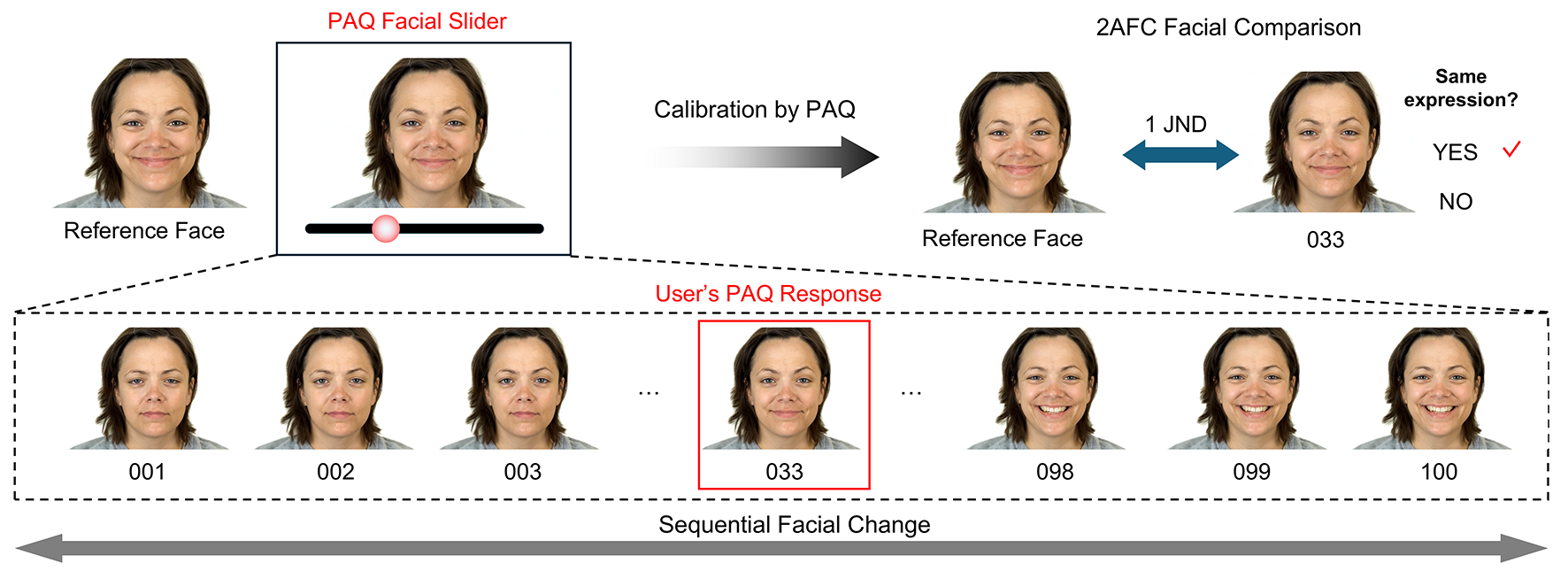}
    \vspace{-18pt} 
    \caption{Illustration of the PAQ-calibrated 2AFC task. First, a PAQ slider response identifies the participants' just noticeable difference with respect to the reference face within a 100-frame continuum. Next,  a paired comparison (2AFC) is elicited between the reference and a 1-JND separated face.}
    \label{fig:ABT_task}
\end{figure}
The concrete example that will form the focus of the rest of this paper is the 2AFC facial comparison task, which presents participants with pairs of fixed face stimuli and asks them to judge whether these two faces share the same facial expression. However, miscalibration can occur in this mechanism  because individuals differ substantially in their perceptual sensitivity to identical facial stimuli. Consequently, any fixed level of emotion in the face stimulus  imposes different levels of cognitive demand across participants. This in turn introduces systematic heterogeneity in task difficulty that confounds behavioral and neural measures. 

\begin{figure}
    \centering
    \includegraphics[width=\linewidth]{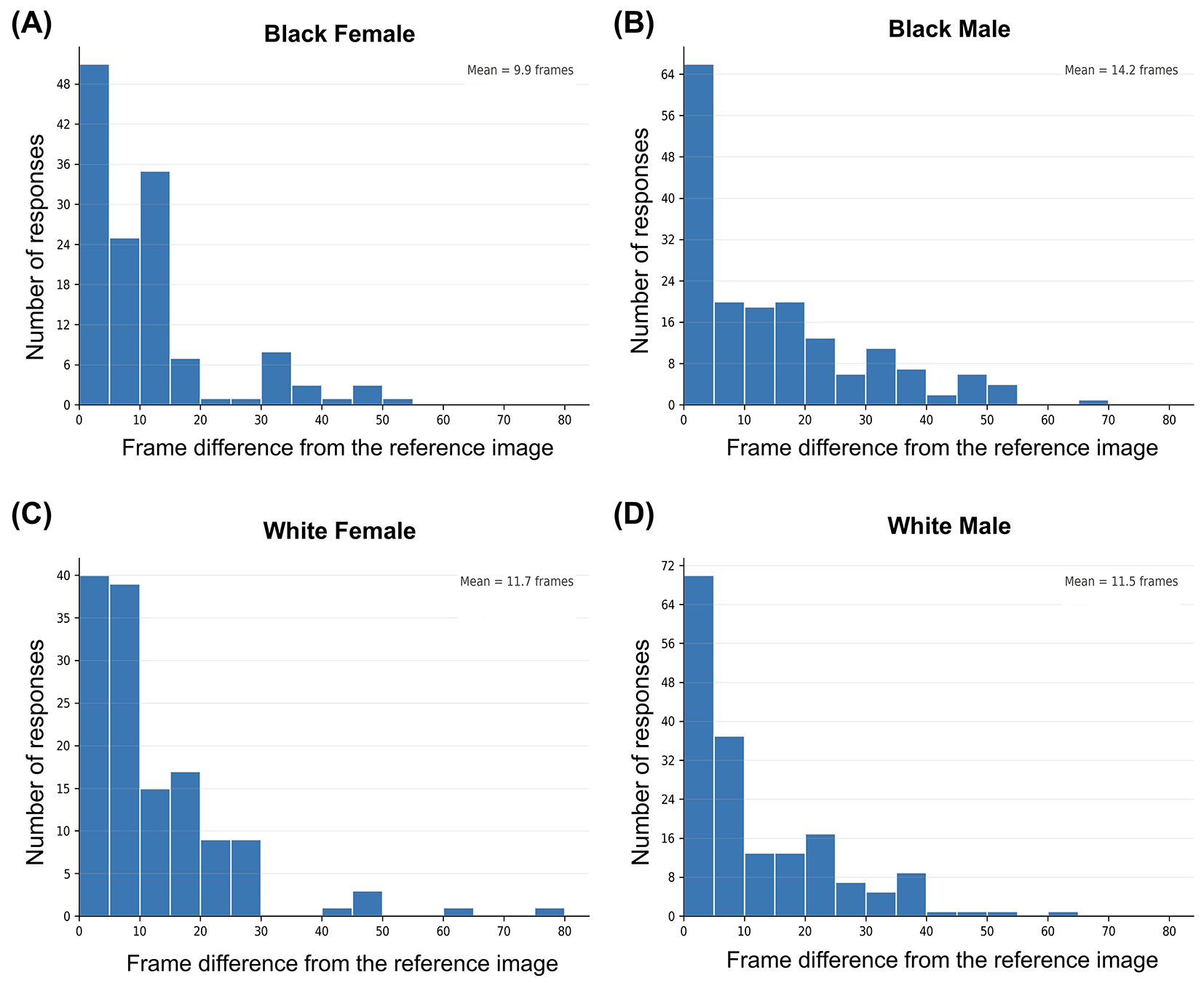}
    \vspace{-18pt} 
    \caption{Distribution of 1-JND among participant responses when they are presented with reference faces from a particular demographic group. The x-axis in all cases measures the absolute frame index difference between two compared stimuli. For each demographic group, JNDs are aggregated across all reference identities from that group. This plot reveals that JNDs vary considerably across individuals and reference identities. More results are in Appendix~\ref{app:jnd_dist}.}
    \label{fig:jnd_distribution}
\end{figure}

To fully address the perceptual miscalibration issue, we introduce a calibration paradigm based on \emph{perceptual adjustment queries} (PAQs)~\cite{xu2023perceptual}. As illustrated in Figure \ref{fig:ABT_task}B, participants complete a PAQ facial slider task, in which a reference face (neutral expression) is presented alongside a slider spanning a sequential continuum of facial changes from neutral to happy/sad (faces 001–100). These continuous PAQ facial stimulus paths are generated by a novel VLM prompting pipeline, which enables fine-grained, artifact-free traversal of the affective spectrum. Each participant is asked to identify the face whose expression is just-noticeably-different from the reference face, i.e. one that falls exactly 1-JND away from the reference (e.g., face 033 in the figure). This response serves as a person-specific calibration of affect sensitivity. In particular, we should expect the task of distinguishing the reference face from that individual's response to be ``self-normalized" for task difficulty. By sampling multiple reference face stimuli along the affect spectrum and asking a PAQ query, we recover each participant's full perceptual scaling profile. In conducting our experiments, we find that JND does indeed vary substantially across individuals: a participant with high emotional sensitivity produces a smaller JND, while one with lower sensitivity produces a larger one. This is indeed borne out in practice (see Figure~\ref{fig:jnd_distribution}) and justifies the need for calibration.  

After collecting PAQ responses, we then measure the extent of calibration achieved. Each face pair, consisting of the reference face and the participant's 1-JND face, is passed into a downstream 2-Alternative Forced Choice (2AFC) facial comparison task, where participants judge whether the two faces have the same expression, and also if the various tasks are of equal difficulty. We find that PAQ calibration successfully normalizes task difficulty both across individuals and across tasks compared to both non-calibrated and population-level calibrated 2AFC task. It yields significant reductions in within-subject and inter-subject variance in participants' metacognition judgment. Taken together, our contributions establish that PAQs can be robustly constructed and used as a principled framework for perceptually grounded affective calibration.

The rest of this paper is organized as follows. Section~\ref{sec:related} discusses related work on existing facial affect paradigms, PAQ for calibration, and controllable facial generation. Section~\ref{sec:method} formalizes the normalization problem and presents the two-step PAQ calibration framework. Section~\ref{sec:exp} 
describes our user interface design and data collection pipeline. 
Section~\ref{sec:results} reports statistical results from our user study, demonstrating that PAQ significantly improves perceptual equality across participants in the downstream 2AFC task.
Section~\ref{sec:conclusion} discusses the limitations as well as potential improvements to our framework.

\section{Related work}
\label{sec:related}

We discuss related work under multiple verticals.
\paragraph{2AFC task and calibration limitations} The 2AFC task~\cite{fechner1860elemente, posner1980orienting} is a foundational psychophysical paradigm in facial affect perception research. It presents pairs of morphed faces at differing emotion intensities and asks participants to judge which expresses the target emotion more strongly, yielding psychometric functions and perceptual boundary estimates grounded in signal detection theory \cite{hautus2021detection}. Its structured trial design and quantitative output have made it the standard tool for measuring perceptual thresholds in emotional expression.
However, the standard 2AFC paradigm suffers from a systematic miscalibration: the physical intensity difference between stimulus pairs is fixed at the design stage and held constant across all participants. This approach ignores the well-documented variability in individual detection thresholds, i.e., perceptual fields differ in location within expression space, and population-level factors including age, sex, and clinical state introduce further heterogeneity. For example, abused children show heightened sensitivity near the anger category boundary~\cite{pollak2002early}, depressed individuals exhibit a generalized deficit in recognizing both positive and negative facial expressions~\cite{surcinelli2006facial}, and individual emotional states are significantly correlated with the interpretation of facial affects~\cite{bae2022correlations}. Because task difficulty is fixed in stimulus space rather than equated in perceptual space, individual perceptual differences become conflated with the affective effects under investigation, undermining the validity of cross-participant comparisons.

\paragraph{PAQ for individual calibration} To normalize task difficulty, we resort to a mechanism that locates each participant's JND boundary along the morph trajectory before downstream analysis. PAQs \cite{xu2023perceptual} serve this purpose: a user adjusts a slider along a continuous stimulus path until the stimulus becomes perceptually equivalent to a reference. A single response implicitly labels the entire traversed path, yielding $O(N^2)$ contrastive supervision signals and outperforming ordinal queries. Chen et al. \cite{10735273} utilized this query mechanism to obtain personalized JND in color perception from continuously varying colors. The affective bias task shares precisely the same structure: face stimuli lie on a continuous morph trajectory as affect spectrum, and a single PAQ recovers each participant's 1-JND point along that trajectory, directly characterizing how they perceptually scale facial expressions. PAQs are therefore a natural and principled instrument for individual-level calibration.

\paragraph{Controllable facial generation} 
PAQs require the generation of a continuous path along affect space.
Generating facial expressions with fine-grained control has been pursued along three main lines. AU-based methods \cite{ekman1978facial, pumarola2018ganimation} conditionally generate faces on anatomically grounded AU intensity vectors. GAN-based disentanglement methods, including StyleGAN \cite{karras2019style}, starGAN \cite{choi2018stargan} and DiscoFaceGAN \cite{deng2020disentangled} enable expression editing via latent space traversal. Diffusion-based approaches \cite{rombach2022high, brooks2023instructpix2pix} offer richer semantic control in facial generation by iteratively denoising from a learned latent space. However, none of these methods guarantee the perceptual smoothness and identity consistency required by a PAQ. In particular, none produce a continuous, monotonic affective trajectory at a resolution sufficient to locate an individual JND boundary, while keeping all non-affective attributes fixed. Our approach is therefore based on vision-language models, which offer a principled interface for semantic control over the affect continuum without committing to a fixed generative architecture.

\section{Methodology}
\label{sec:method}

\begin{figure}
    \centering
    \includegraphics[width=\linewidth]{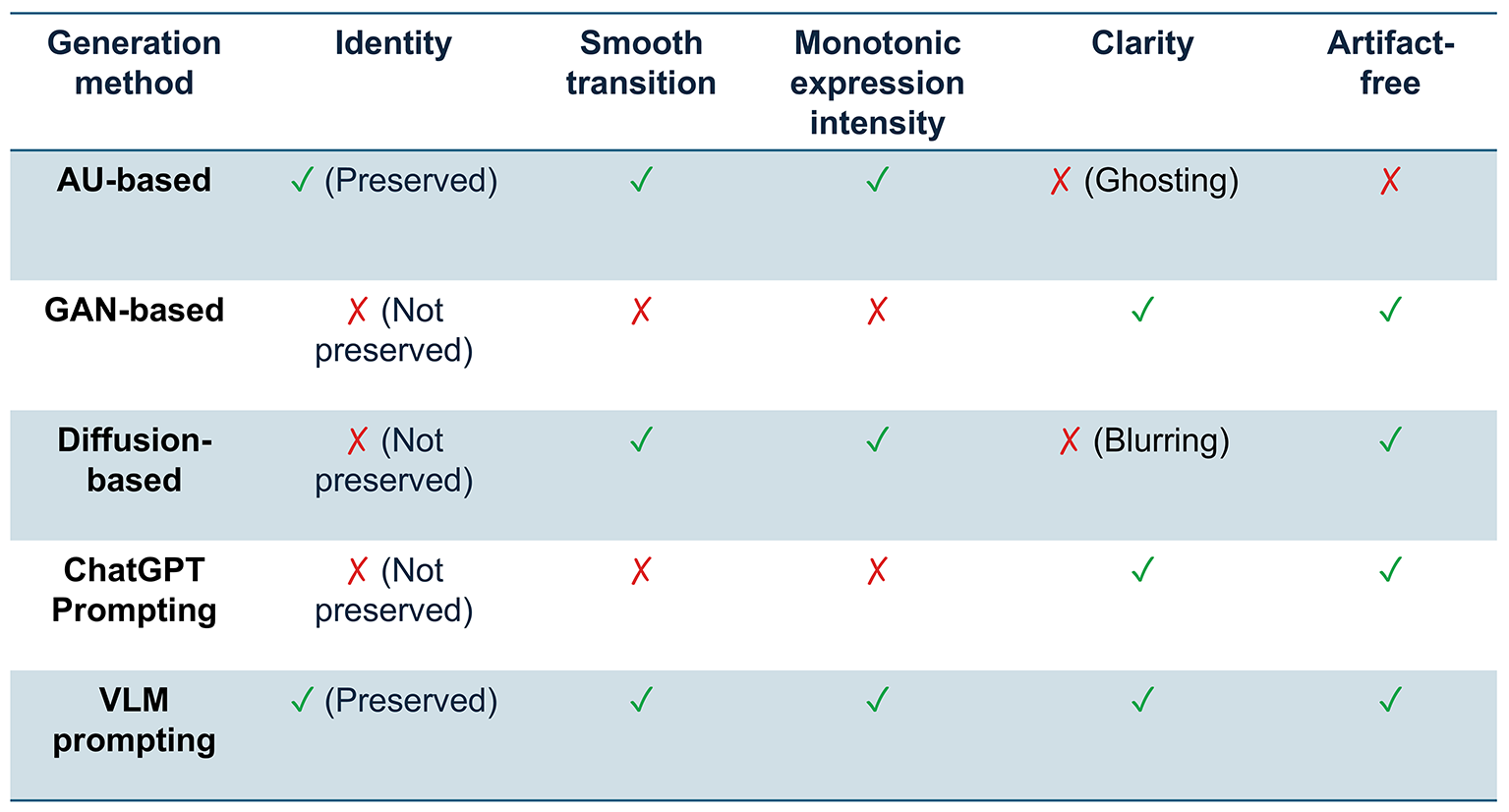}
    \vspace{-18pt} 
    \caption{Qualitative Comparison of facial expression generation methods (including AU-based generation, GAN-based generation, Diffusion-based generation, ChatGPT prompting and VLM prompting) across five criteria: identity preservation, smooth transition, monotonic expression intensity, clarity, and absence of artifacts.}
    \label{fig:face_gen_comp}
\end{figure}

\subsection{Problem formulation}
Let $x \in \{0, \ldots, 100\}$ denote the physical morph intensity along the affect spectrum, where $x = 0$ is neutral and $x = 100$ is maximally valenced (e.g., fully happy or sad). Each participant maps physical intensity to perceived emotional strength via a latent monotone perceptual function. Because this function varies across individuals, identical physical stimuli evoke different magnitudes of perceived emotional change. The same morph step $\Delta x$ may be clearly noticeable to one participant yet imperceptible to another. Raw behavioral responses are therefore inaccurate across participants without prior perceptual calibration. In addition to variation at the individual level, the perceptual function may additionally vary over the 
affect spectrum.

The goal of calibration is to re-express the stimulus distances in the 2AFC task as units of each individual's perceptual scale (i.e., that individual's JND) so that every participant is always judging a stimulus pair that sits at the boundary of their own discriminability, making perceived task difficulty uniform across individuals by task reconstruction. Let $\delta_{i}(x)$ denote participant $i$'s individual JND with respect to reference affect $x$.

\subsection{Generating PAQ paths and tasks}\label{sec:method_paq}

\begin{figure}
    \centering
    \includegraphics[width=\linewidth]{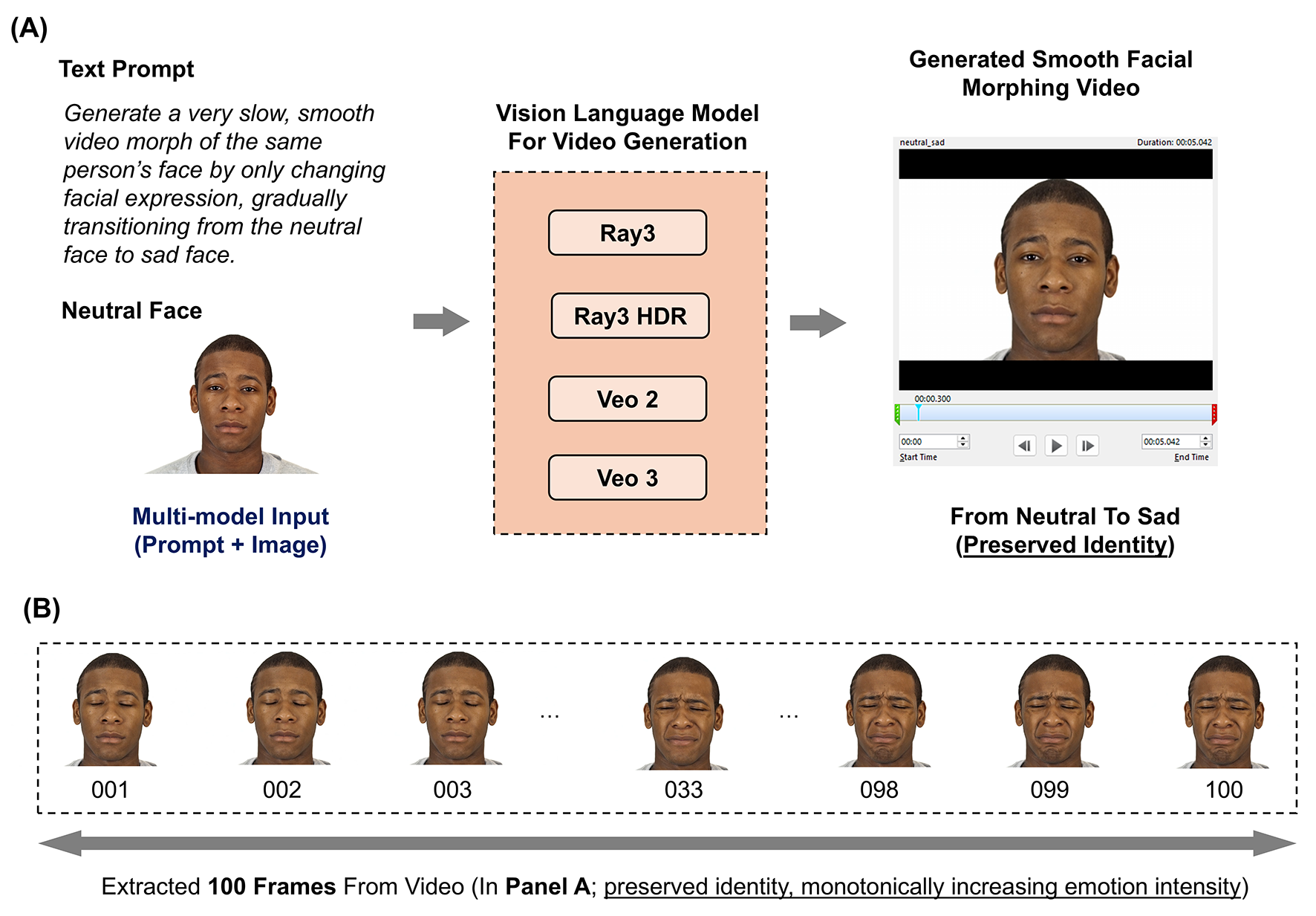}
    \vspace{-18pt} 
    \caption{VLM facial PAQ Path Generation Pipeline: (A) A neutral face serves as the first frame, paired with a text prompt instructing the model to generate a smooth video morph of the same face transitioning from neutral to a target expression (happy or sad). Candidate VLM models are Ray3, Ray3 HDR, Veo2, and Veo3. (B) The generated video is post-processed by extracting individual frames to construct a 100-frame PAQ facial morphing sequence.}
    \label{fig:vlm_gen}
\end{figure}

\paragraph{Facial PAQ path generation} 
A PAQ requires a stimulus path that is continuous, monotonically ordered, identity-preserving, and dense enough to locate an individual JND boundary.
Generating such a path is challenging from a technical standpoint. For example, AU-based methods preserve identity but suffer from ghosting and artifacts; GAN-based methods produce clear, artifact-free outputs but fail to preserve identity or ensure smooth, monotonic expression transitions; diffusion-based methods offer smooth transitions but introduce blurring in the middle and lose identity at the beginning; ChatGPT prompting lacks smoothness and monotonicity in expression intensity during morphing as well as identity preservation. See Figure~\ref{fig:face_gen_comp} for an overall comparison and Appendix~\ref{app:face_gen} for more details.

We overcome these challenges by building on a vision-language generative model (VLM) as a unified multimodal framework for semantic control over the affect continuum (see Figure~\ref{fig:vlm_gen}). Concretely, we prompt the VLM with a structured instruction that enforces identity preservation and expression monotonicity simultaneously as generation constraints: \texttt{Generate a very slow, smooth video morph of the same person’s face by only changing facial expression, gradually transitioning from the neutral face in the first frame to a happy (resp. sad) face in the last frame.}  This multimodal formulation, combining a visual identity anchor (a neutral face) with a linguistic trajectory specification, instructs the model to modify only expression-related action units while holding identity-specific attributes (such as hairstyle or facial geometry) fixed. The key advantage over AU-based or GAN-based alternatives is that the VLM operates directly in semantic space, decoupling the expression trajectory from the generative architecture and avoiding the identity drift and non-monotonic perceptual changes that afflict latent-space traversal methods. 
Our text prompt and reference face are fed to four candidate VLMs Ray3, Ray3 HDR, Veo 2 and Veo 3, and we select (by-eye) the highest-quality facial morphing video from all candidate models.  From each synthesized video, we extract 100 sequential frames at uniform intervals, forming a monotonically ordered discrete sequence from neutral to peak expression that constitutes the PAQ stimulus path. We examine each path to ensure stimuli are artifact-free. Crucially, equal spacing in frame index does not imply equal spacing in perceptual affect, which is precisely why individual calibration is necessary to locate each participant's JND within the sequence. 

\paragraph{Reference face sampling}
Face identities are drawn from the Chicago Face Database~\cite{ma2015chicago}, which provides public, high-resolution, demographically diverse facial images spanning eight groups: Black female, Black male, White female, White male, Indian 
female, Indian male, Malaysian female, and Malaysian male. For each demographic 
group, one identity is selected with its corresponding 
neutral, sad, and happy expressions, yielding 8 reference 
identities in total. Each identity's neutral-to-sad and 
neutral-to-happy expression pairs serve as input to the 
VLM-based pipeline, which generates a continuous 100-frame 
PAQ facial path along each valence direction. Reference 
stimuli along the affect spectrum are then constructed by 
uniformly sampling two anchor points per identity from each 
generated path, ensuring coverage across the full sad-to-happy 
continuum. The aggregated reference stimulus set shared across 
all participants is: $\{\text{sad}_{88}, \ldots, \text{sad}_{11},
\text{neutral}, \text{happy}_{11},\ldots, 
\text{happy}_{88}\}$, where subscripts denote frame indices 
along the 100-frame PAQ path. This stimulus set of reference faces is common across all individual participants, ensuring that any 
observed differences in behavioral responses reflect 
individual perceptual variation rather than differences 
in stimulus assignment.

\paragraph{The PAQ task} With the stimulus path established, at each querying trial, a reference facial stimulus is presented at a fixed but undisclosed position along the affect spectrum. The participant adjusts a continuous slider until the comparison face first appears to affectively different from the reference. The distance between the reference and the participant $i$'s stopping point defines their JND at that reference face $\delta_i(x)$, the smallest perceptual step that registers as a distinguishable change in emotional expression. Repeating this procedure with multiple reference facial stimuli yields an informative set of JND measurements along the full affect continuum.

\subsection{Calibrating a downsteam 2AFC task}

As our downstream task, we employ a 2AFC same-different paradigm in which participants judge whether two simultaneously presented faces express the same or different emotional strength. This serves as a testbed to measure  whether we have achieved the desired level of calibration. While the baseline would present stimulus pairs as a fixed physical inter-stimulus distance, we now use each participant's PAQ-derived JNDs $\delta_{i}(x)$ to reconstruct a perceptually calibrated trial set, where each time, the pair of faces presented to participant $i$ are now given by $x$ and $x + \delta_i(x)$. In words, two faces in each 2AFC trial are placed at exactly 1-JND separation along the affect spectrum, positioned at the left and right boundaries of that participant's individual JND interval.

\subsection{Response time and metacognition on downstream task} To verify whether PAQ-based calibration successfully equalized task difficulty, we collect two complementary behavioral measures. First, response time (RT) is recorded on every trial as the interval from stimulus onset to keypress. We expect that after calibration, RTs for the same 2AFC task should be smaller and less variable across participants as well as for the same participant for different 2AFC tasks. Second, after every block of four consecutive trials, participants provide a binary metacognitive judgment indicating whether these four comparisons were equally difficult. This metacognitive signal provides a direct, subjective verification of difficulty equalization. Together, the two measures form the empirical basis for testing the hypothesis that PAQ-derived JND normalization produces more uniform perceived task difficulty than the uncalibrated baseline. We also compare with population-level Weibull calibration, which is described next.

\section{Experiments}
\label{sec:exp}

\subsection{Baseline and calibration strategies}
For all of our experiments, two calibration strategies: population-level calibration and PAQ-based calibration, are compared against the non-calibrated 2AFC baseline. In the non-calibrated baseline, stimulus pairs are drawn from fixed frame intervals $\Delta x = 11$ (frame indices) and we ask participants whether the faces share the same expression. The \emph{population-level} calibration first collects binary ``same or different'' responses from all $N$ participants in a non-calibrated 2AFC task at a fixed inter-stimulus distance $\Delta x = 11$ (frame indices). Responses are aggregated across participants to obtain the empirical discrimination frequency $\hat{p}(\Delta x)$, which is then fit using a Weibull CDF: $\hat{p}(\Delta x) = 1 - e^{-(\Delta x / \hat{\lambda})^2},$ from which the population-level scaling parameter $\hat{\lambda}$ is inferred. 
Having inferred $\hat p$, we may now ``infer" a population-level JND. Following prior work~\cite{torgerson1958theory}, we set $\Delta x^\star$ to be the value of $\Delta x$ for which $\hat p = 0.5$, which is given by $\hat{\lambda} \sqrt{\ln 2}$. While principled, this approach collapses individual differences into a single, shared threshold. In contrast to population-level calibration, the \emph{PAQ-based calibration} method described above bypasses group-level model fitting entirely by directly using each participant's JND at each reference point through personal perceptual adjustment. In particular, we do not recover a global $\Delta x^*$, but individual-level $\Delta x^*_i(x) = \delta_i(x)$ for each participant $i$ and reference point $x$. 

\subsection{Crowdsourcing setup and user interface}

Participants were recruited via 
Prolific and compensated at a standard hourly rate consistent with 
Prolific's fair pay guidelines (\$8/hour), determined based 
on estimated task completion time. Participants were required to have normal or corrected-to-normal vision and no prior exposure to stimuli. Participants were instructed that the study involved judging facial 
expressions and that their response times and binary responses would be recorded. They were informed that participation was voluntary, that they could 
withdraw at any time without penalty, and that their data would be 
anonymous prior to analysis. No instructions were given that are likely 
to have influenced the study findings. All participants provided written 
informed consent prior to participation. The experimental interface is implemented as a web-based application, administered on a laptop or desktop at 100\% zoom. The interface comprises three sequential stages: 

\textbf{Stage 1 (Non-calibrated 2AFC and population-level calibrated 2AFC)} presents pairwise face comparisons at a fixed physical inter-stimulus distance $\Delta x$ shared uniformly across all participants: $\Delta x = 11$ frame indices for the non-calibrated baseline, and $\Delta x^\star = \hat{\lambda}(\log 2)^{1/2}$ for the population-level calibration condition, where $\hat{\lambda}$ is estimated by fitting a Weibull function to the aggregate binary responses collected in the pilot study. In each trial, the left face serves as the reference anchor, uniformly sampled along the affect spectrum, with the right face placed at exactly $\Delta x$ (or $\Delta x^*$) frame indices away. Participants judge whether the two faces express the same or different emotional strength via keyboard press (Y for Same, N for Different), responding as quickly and accurately as possible. Response time is recorded from stimulus onset to keypress. Face placement is randomized across trials to minimize positional bias. \textbf{Stage 2 (PAQ calibration)} implements an expression matching task in which a reference face displaying a target emotional expression (e.g., happy or sad) is presented on the left as a fixed anchor, while a test face of the same identity starts as neutral on the right. The participant controls a horizontal slider that continuously morphs the test face the target expression. The participant is instructed to stop at the earliest point where the two faces appear affectively equivalent. This task is not timed to encourage careful perceptual judgment, and the slider position at response is recorded as the participant's 1-JND estimate for that identity and expression combination (see Figure~\ref{fig:ABT_task}B). \textbf{Stage 3 (PAQ calibrated 2AFC)} uses each participant's PAQ-derived JNDs to reconstruct and present stimulus pairs at exactly that separation. Each trial begins with a fixation cross to orient spatial attention, followed by simultaneous presentation of two face images side-by-side. Similarly to Stage 1, participants judge whether the two faces have the same or different intensity of emotion and the response time for each random trial is recorded (see Figure~\ref{fig:ABT_task}B).

\subsection{Data collection pipeline}
\textbf{PAQ vs. Non-calibration study ($N = 52$)}: The pilot study collects data across three sequential stages: an non-calibrated 2AFC task at fixed inter-stimulus distance $\Delta x = 11$, followed by PAQ calibration to elicit each participant's individual JNDs, followed by a PAQ-calibrated 2AFC task. In both 2AFC stages, response time (RT) and per-block binary metacognitive difficulty judgments are recorded as the primary behavioral indicators. The aggregate binary responses from the non-calibrated 2AFC stage are used to fit the population-level Weibull function and estimate the shared threshold $\hat{\lambda}$, which serves as the population-level calibration baseline in the main study, which is described next. \textbf{PAQ vs. Population-level calibration study ($N = 72$)}: The main study recruits a new cohort matched to the pilot in demographic composition — balanced across age, sex, and demographic groups. Participants complete three sequential stages: a population-level calibrated 2AFC task using $\Delta x^* = \hat{\lambda}(\log 2)^{1/2}$ estimated from the pilot, followed by PAQ calibration, followed by a PAQ-calibrated 2AFC task. RT and metacognitive difficulty judgments are collected in both 2AFC stages, enabling direct comparison between population-level and PAQ calibration among the same participants.

\section{Results and Statistical Analysis} 
\label{sec:results}

\subsection{Results from metacognitive judgements}
Table~\ref{tab:metacog} reports the paired metacognitive difficulty judgments, comparing the non-calibrated 2AFC task against PAQ calibration. Among the discordant pairs in the pilot study, 47 
blocks improved against 19 worsening, yielding an improvement rate of $71.2\%$. Among the discordant pairs in the main study, 52 blocks transitioned from not-equal to equal 
perceived difficulty following PAQ calibration (improvement), while 25 transitioned 
in the opposite direction (worsening). This yields an improvement rate of $67.5\%$ among examples in which a change in difficulty was perceived.  In sum, PAQ not only calibrates effectively against the naive baseline, but also maintains a 
significant improvement rate even when compared against a principled, 
population-level calibration technique. 


\begin{table}
\centering
\begin{tabular}{|c|c|c|c|c|}
\hline
\multicolumn{3}{|c|}{\textbf{PAQ \textit{vs.} (Baseline and Pop-Level Calibration)}} & \multicolumn{2}{c|}{\textbf{PAQ}} \\
\cline{4-5}
\multicolumn{3}{|c|}{} & Equal & Not Equal \\
\hline
\multirow{2}{*}{\textcolor{blue}{\textbf{Group I}}} & \multirow{2}{*}{\textbf{Baseline}} & Equal & 104 & 19 \\
\cline{3-5}
 & & Not Equal & 47 & 38 \\
\hline
\multirow{2}{*}{\textcolor{blue}{\textbf{Group II}}} & \multirow{2}{*}{\textbf{\shortstack{Pop-Level \\ Calibration}}} & Equal & 128 & 25 \\
 & & Not Equal & 52 & 83 \\
\hline
\end{tabular}
\vspace{2pt} 
\caption{Table of paired metacognitive difficulty judgments between baseline/population level calibration and PAQ calibration. }
\label{tab:metacog}
\end{table}



We examine these improvements in more detail via statistical analysis, using a one-sided exact McNemar test.
For each comparison, the null hypothesis asserts that PAQ calibration has no systematic 
effect on perceived task difficulty equalization, i.e., transitions from unequal to 
equal difficulty (improvement) and from equal to unequal difficulty (worsening) are equally 
likely among discordant blocks: $H_0 : P(c) = P(b) = 0.5$. The alternative hypothesis is directional, asserting that improvement is strictly 
more likely than worsening following PAQ calibration: $H_1 : P(c) > P(b)$. Under $H_0$, the number of improving discordant pairs $c$ follows a binomial distribution: $c \mid H_0 \sim \text{Binomial}(c + b,\ 0.5)$. The test statistic is $c$, the observed count of blocks transitioning 
from not-equal to equal difficulty. The one-sided $p$-value is the probability of observing 
$c$ or more improving pairs under $H_0$:$p = P(X \geq c \mid X \sim \text{Binomial}(c + b,\ 0.5))$. \textbf{Baseline vs PAQ:}
The discordant pairs are $c = 47$ and $b = 19$, yielding $c + b = 66$ total discordant 
blocks. Under $H_0$: $    p = P(X \geq 47 \mid X \sim \text{Binomial}(66,\ 0.5)) = 0.0004.$ Since $p = 0.0004 < \alpha = 0.05$, we reject $H_0$. The improvement rate among 
discordant pairs is $47/66 = 71.2\%$, significantly exceeding the chance level of $50\%$. 
We conclude that PAQ calibration significantly increases the proportion of blocks perceived 
as equally difficult relative to the non-calibrated pre-PAQ baseline. \textbf{Pop-level vs PAQ:}
The discordant pairs are $c = 52$ and $b = 25$, yielding $c + b = 77$ total discordant 
blocks. Under $H_0$: $ p = P(X \geq 52 \mid X \sim \text{Binomial}(77,\ 0.5)) = 0.0014.$ Since $p = 0.0014 < \alpha = 0.05$, we reject $H_0$. The improvement rate among 
discordant pairs is $52/77 = 67.5\%$, significantly exceeding chance. We conclude that PAQ 
calibration significantly improves perceived task difficulty equalization over the 
fixed-distance population-level baseline condition. Taken together, the rejection of $H_0$ in both tests provides strong statistical 
evidence that PAQ-based individualized calibration produces a systematic and significant 
improvement in the equalization of perceived task difficulty, both over the absence of 
any calibration and the population-level psychometric baseline.

In addition, to respect the within-participant correlation structure,
we first collapse each participant's four blocks into a single
\emph{net-shift score} and reduce
this to the sign of the net shift, so that every participant enters the
analysis exactly once. We then apply McNemar's exact one-sided test, which counts only the
discordant participants and tests them against $\text{Binomial}(c + b,\, 0.5)$:
under $H_0$ of no systematic effect, a discordant participant is equally
likely to shift in either direction, so the favorable count $c$ follows
$\text{Binomial}(c + b, 0.5)$ and the one-sided $p$-value is
$P(X \geq c)$. \textbf{Baseline vs.\ PAQ:} the discordant participants split $c = 25$
favoring PAQ against $b = 5$ favoring baseline, giving $c + b = 30$ total
discordant participants. Under $H_0$,
$p = P\!\left(X \geq 25 \mid X \sim \text{Binomial}(30, 0.5)\right) = 0.0002$,
so PAQ calibration significantly increases the proportion of confident
(``equal'') meta-cognitive judgments relative to the non-calibrated
baseline. \textbf{Pop-level vs.\ PAQ:} the discordant participants split $c = 29$
favoring PAQ against $b = 8$ favoring the population-level calibration,
giving $c + b = 37$ total discordant participants. Under $H_0$,
$p = P\!\left(X \geq 29 \mid X \sim \text{Binomial}(37, 0.5)\right) = 0.0004$.

\subsection{Results from response time}

\begin{figure}
    \centering
    \includegraphics[width=\linewidth]{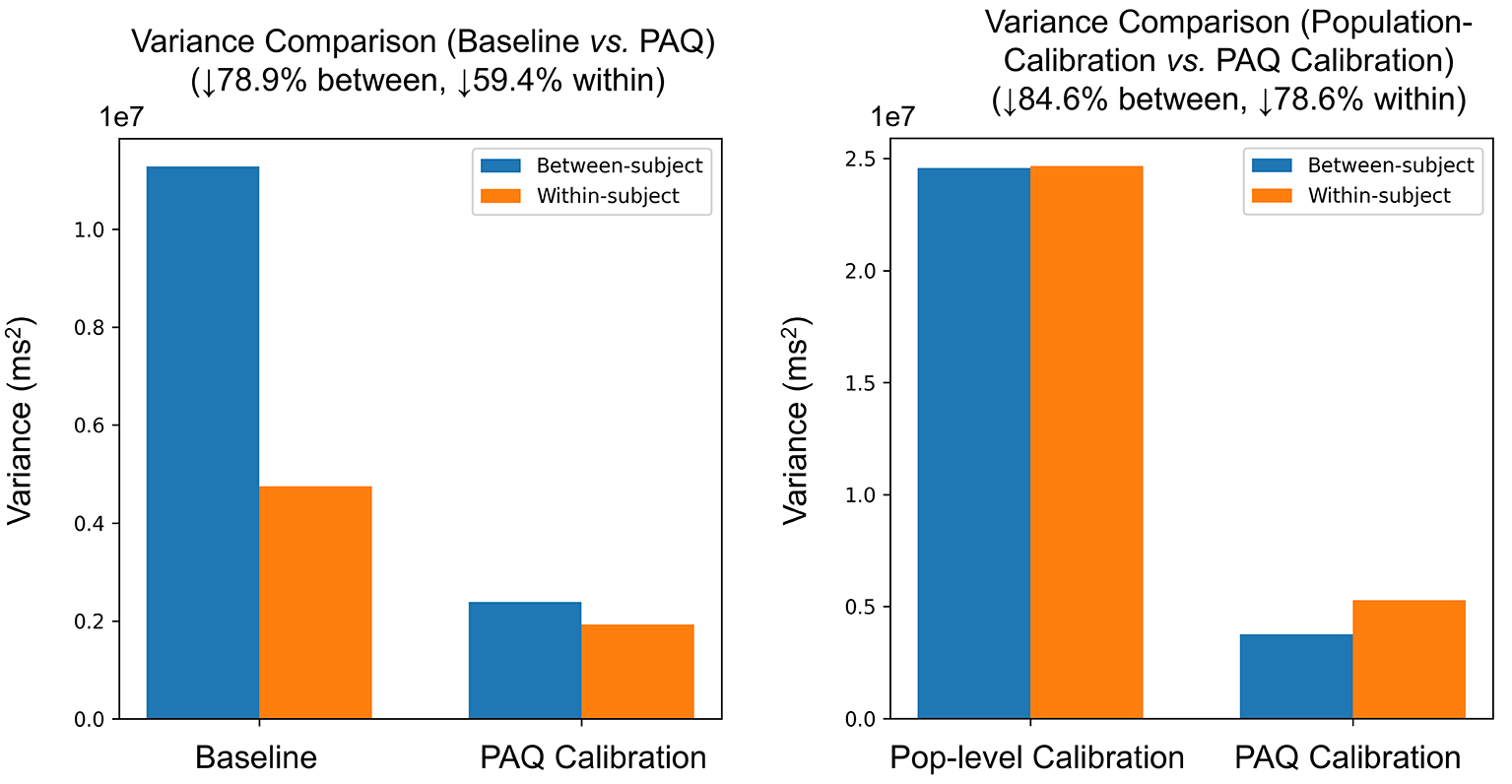}
    \vspace{-18pt} 
    \caption{Variance of response time before and after PAQ calibration. Each panel shows between-subject variance (blue) and within-subject variance (orange) under two calibration conditions. Left: Comparing no calibration (Baseline) against PAQ calibration. Right: Comparing population-level calibration against PAQ calibration.}
    \label{fig:variance_components}
\end{figure}

\begin{figure}
    \centering
    \includegraphics[width=\linewidth]{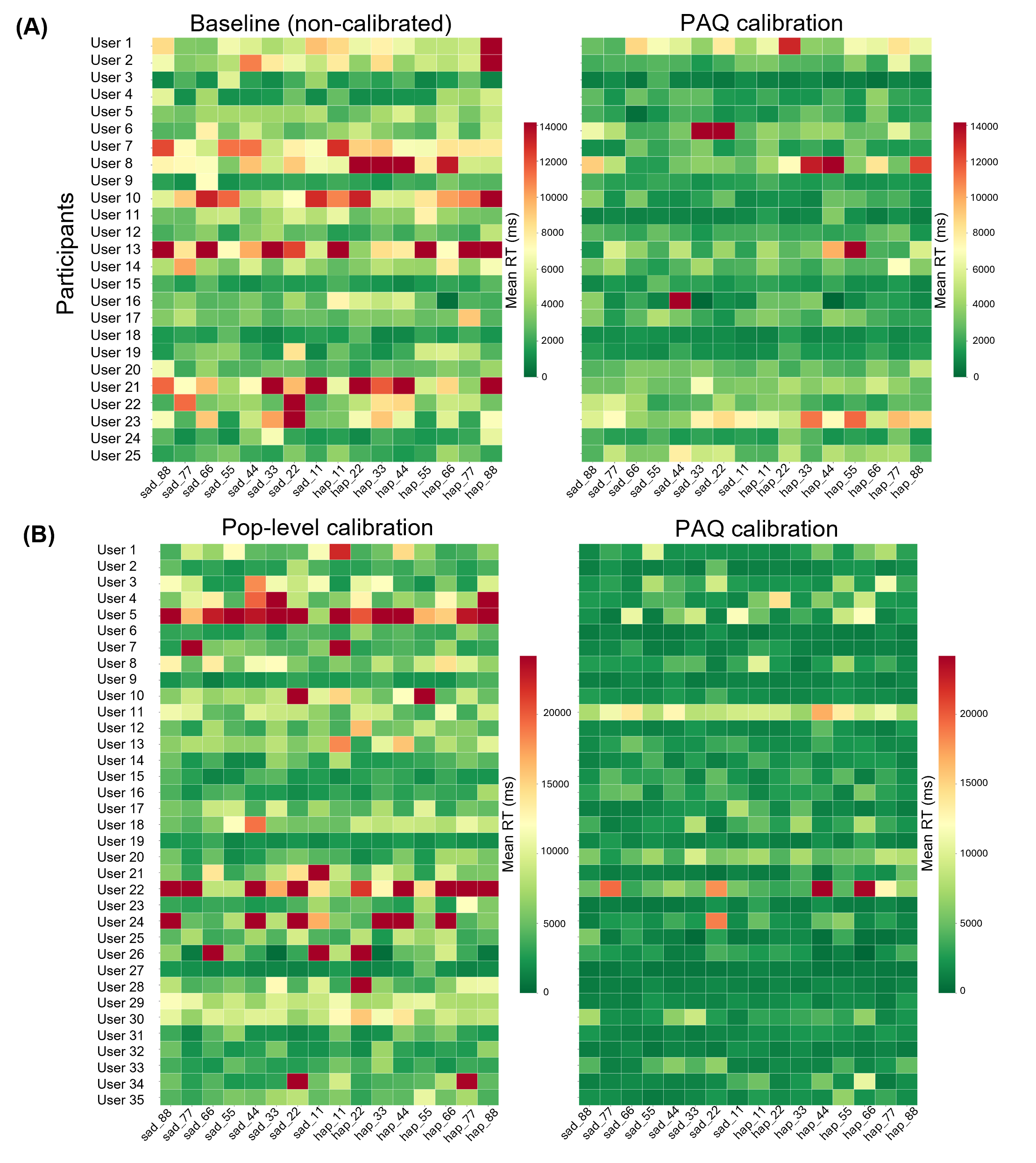}
    \vspace{-18pt} 
    \caption{(A): Side-by-side heatmap comparisons of response times (RT) across the affect spectrum (non-calibrated baseline vs. PAQ calibration) on $25$ randomly sampled participants. Left panel: Mean RT from the non-calibrated baseline model, where the mean is taken over responses to the different reference identities. Right panel: Mean RT after applying PAQ calibration. Each row represents a participant, and each column represents a position along the affect spectrum from sad to happy. Color intensity indicates mean RT in milliseconds (green = faster responses, red = slower responses). Similar in (B) on $35$ participants with population-level calibration (left panel) vs. PAQ calibration (right panel).}
    \label{fig:heatmap_25}
\end{figure}

Figure~\ref{fig:variance_components} decomposes total RT variability into two orthogonal sources: between-subject variance (the spread of individual mean RTs across participants) and within-subject variance (the average spread of RTs across stimulus positions within a single participant). It tracks how each changes from baseline in the left panel (population-level calibration in the right panel) to PAQ calibration. \textbf{Baseline vs. PAQ:} Before PAQ, between-subject variance dominates at approximately $11.3\text{ s}^2$ and within-subject variance at $\sim 4.7 \text{ s}^{2}$. After PAQ calibration, both variances drop sharply, but the reductions are asymmetric: between-subject variance contracts by 78.9\% to $\sim2.4\text{ s}^{2}$, while within-subject variance falls by 59.4\% to $\sim2.0\text{ s}^{2}$, so the two components become nearly equal in magnitude. \textbf{Pop-level vs. PAQ:} Before PAQ calibration, both between-subject variance and within-subject are at approximately $24.7\text{ s}^2$. After PAQ calibration, both variances drop sharply, but the reductions are asymmetric: between-subject variance contracts by $84.6\%$ to $\sim 3.8\text{ s}^2$, while within-subject variance falls by $78.6\%$ to $\sim 5.3\text{ s}^2$, so the within-subject component becomes modestly larger than the between-subject component post-calibration. Overall results show that PAQ calibration improves RT data quality through two complementary mechanisms: homogenizing the participant pool (between-subject component) and, to a lesser extent, stabilizing each individual's moment-to-moment processing (within-subject component).

The affect spectrum heatmaps (Figure~\ref{fig:heatmap_25}A) display 25 sub-sampled participants' mean RT across 16 face-morphing positions ordered from extreme sad ($\text{sad}_{88}$) to extreme happy ($\text{happy}_{88}$), with rows representing individual participants and color intensity encoding RT magnitude (red = slow, green = fast). Without calibration, the Baseline heatmap reveals pronounced between-participant heterogeneity: most participants exhibit uniformly moderate RTs ($\sim$1.5-5 s), but a subset of participants show bright red streaks at one or more spectrum positions, with peak RTs exceeding 40 s, indicating high idiosyncratic variability in how individuals attend to emotionally ambiguous face morphs. After PAQ calibration, the heatmap shows a markedly compressed color range (peak RT reduced to $\sim$25 s) and a more uniformly green matrix, indicating that personalized calibration substantially reduced both the absolute level and the dispersion of RTs across the spectrum. Similarly, the affect spectrum heatmaps (Figure~\ref{fig:heatmap_25}B) display a subsampled cohort's ($N = 35$) mean RT to compare population-level calibration against PAQ calibration. Even with population-level calibration, the heatmap still shows notable between-participant heterogeneity: extreme baseline outliers are partly suppressed, but some participants (e.g., Users 5, 22, 24) retain red streaks across multiple spectrum positions, with peak RTs above $40\,$s, showing that a single shared threshold $\hat{\lambda}$ cannot fit participants whose sensitivity deviates sharply from the population mean. After PAQ calibration, the heatmap turns greener and more uniform, with peak RT dropping to approximately $20\,$s and red streaks largely gone. Remaining red cells (e.g., User 22) are isolated rather than row-wide, suggesting PAQ removes the systematic, identity-wide RT inflation from population-level mismatch, leaving only genuine trial-level uncertainty. The gap between the two calibrated panels is smaller than between baseline and PAQ, confirming population-level calibration is a useful but incomplete fix, while PAQ's individualized normalization offers a further, qualitatively distinct improvement.

\section{Conclusion and discussion}
\label{sec:conclusion}

This work addresses perceptual miscalibration in facial affect tasks, where individual differences in sensitivity and response scaling render raw ratings incomparable across participants. We propose PAQ, a lightweight query mechanism that recovers each participant's JND along the affect spectrum and uses it to replace fixed physical inter-stimulus distances with individually normalized ones in a 2AFC facial comparison task. Across 124 participants, PAQ significantly increased the proportion of trials perceived as equally difficult, outperforming both an uncalibrated baseline and population-level calibration, demonstrating that individual thresholds cannot be substituted by population-level normalization alone. Along the way, we tackle the challenge of constructing a continuous PAQ stimulus path via a novel VLM-based prompting pipeline that synthesizes identity-preserved facial expressions varying monotonically in emotional intensity. An important limitation concerns the ordering of experimental stages. Because participants complete PAQ calibration before the post-PAQ 2AFC task, they become more familiar with the stimuli. In further work, one can address this ordering issue by putting the calibration stage before both baseline and PAQ-calibrated 2AFC tasks. To assess the test/retest reliability of JND derived from PAQ, one can conduct a longitudinal study or a within-visit procedure that mitigates memory effects.

\newpage
\section*{Ethical Impact Statement}

\subsection*{Human Subjects Research}
This experimental protocol was approved by the Institutional Review Board (IRB) 
with the approval number: IRB2025-1286. Please see the Section \ref{sec:exp} for other information including instructions, informed consent, and compensation.

\subsection*{Potential Negative Impacts and Mitigation}

\paragraph{Potential for Deception}
This research does not involve deception. Participants were fully informed 
of full content of the tasks prior to participation. The PAQ calibration 
and 2AFC tasks involve no misleading elements.

\paragraph{Bias and Discrimination} We acknowledge the potential for demographic bias in facial affect 
perception research. To mitigate this, reference face identities were 
deliberately selected from a demographically diverse set spanning eight 
groups (Black female, Black male, White female, White male, Indian female, 
Indian male, Malaysian female, and Malaysian male). Participant recruitment 
was similarly balanced across demographic groups. 

\paragraph{Misuse and Surveillance}
The PAQ framework estimates individualized perceptual thresholds for 
facial affect recognition. While designed for psychophysical 
calibration in research and clinical contexts, we acknowledge that 
perceptual profiling technology could in principle be misused in 
surveillance or affective monitoring systems without individuals' 
informed consent. We strongly discourage such applications and 
advocate for transparent, consent-based use of any perceptual 
calibration methodology derived from this work.

\subsection*{Limits of Generalizability}

\paragraph{Assumptions and Robustness}
Our framework assumes that each participant's perceptual function along the affect spectrum is monotone and well-approximated by a Weibull 
cumulative distribution. Due to fatigue, the monotonicity can be violated, could 
reduce the reliability of PAQ-derived JND estimates. Nevertheless, the number of trials are carefully adjusted to minimize participant fatigue to maintain the accuracy of JND estimation.

\paragraph{Scope of Claims}
Our user studies were conducted on a sample of $N = 124$ participants recruited via Prolific, which, while demographically balanced by design, may not fully represent the global population. In particular, our stimulus set is limited to eight demographic groups, and results may not 
generalize to facial identities or participant populations outside this set. The scalability of PAQ calibration to larger, more diverse cohorts including clinical populations such as those with depression or autism spectrum disorder remains to be demonstrated.

\paragraph{Factors Influencing Task Performance}
Several contextual and individual factors may influence the performance 
of the PAQ calibration and downstream 2AFC task beyond perceptual 
sensitivity. Participants' transient affective states at the 
time of testing including mood, stress, and fatigue, are known to 
shift the interpretation of facial expressions. Second, cognitive factors such as working memory capacity and attention may affect the response precision during the task, as careful perceptual judgment requires sustained attention over the course of the calibration session.

\paragraph{Contextual Sensitivity}
The PAQ framework was developed and validated specifically for facial affect perception along a sad-to-happy continuum. Its extension to other affective dimensions (e.g., arousal, dominance), other emotion 
categories, or cross-cultural settings where the internal representation 
of neutral and emotional expressions may differ substantially.

\section{Acknowledgment}
This work was supported in part by the National Center for Advancing Translational Sciences of the National Institutes of Health under Award Number UL1TR002378 and KL2TR002381, the National Science Foundation award CCF-2107455, and a Google Research Scholar Award. We thank Vivek Anand and Chris Rozell for helpful discussions.

\bibliographystyle{IEEEtran}
\bibliography{references}

\clearpage
\appendices

\section{Additional experimental results}
\subsection{Results in JND distribution}\label{app:jnd_dist}
\begin{figure}
    \centering
    \includegraphics[width=\linewidth]{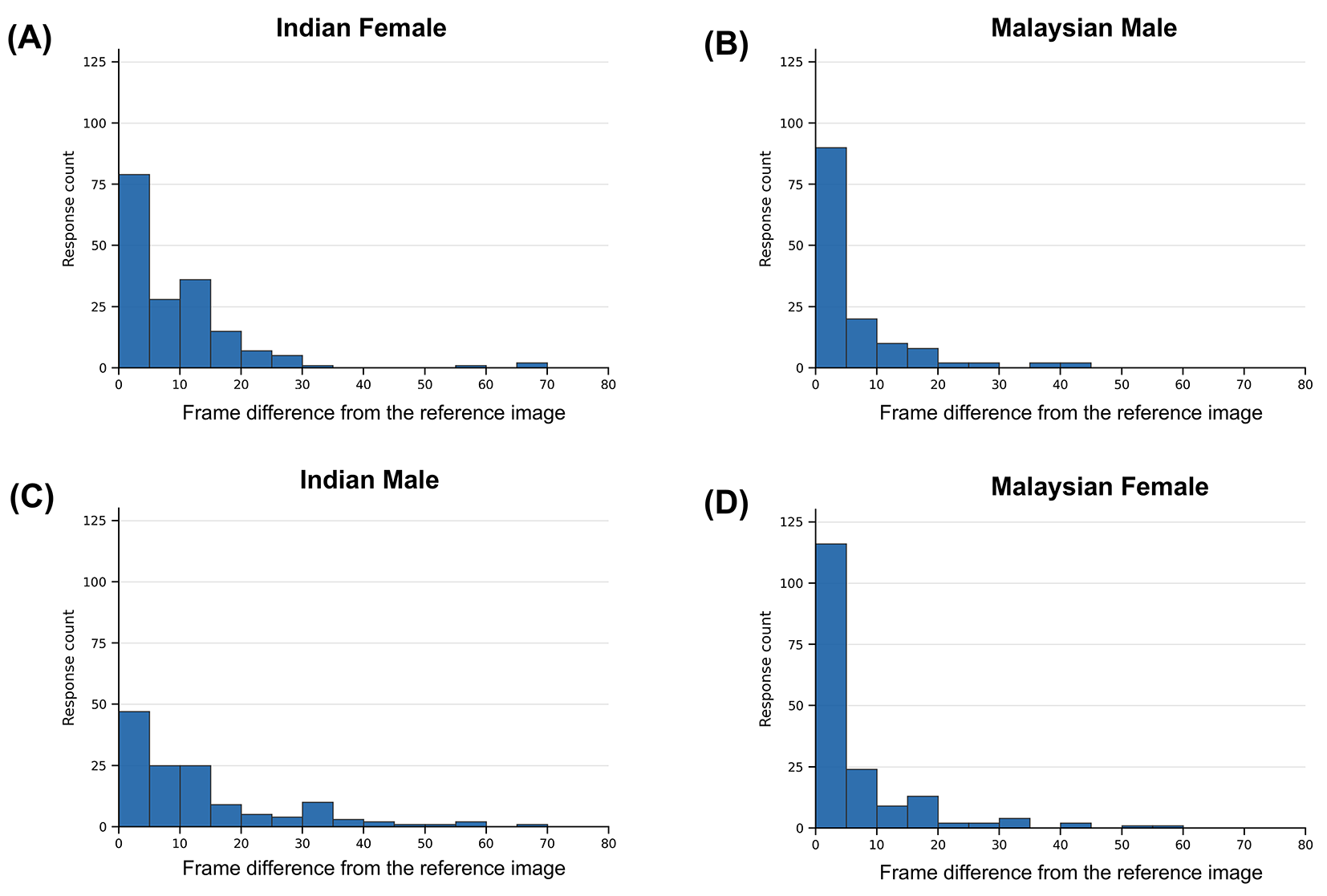}
    \vspace{-18pt}
    \caption{Distribution of 1-JND among participant responses when they are presented with reference faces from a particular demographic group. The x-axis in all cases measures the absolute frame index difference between two compared stimuli. For each demographic group, JNDs are aggregated across all reference identities from that group. This plot reveals that JNDs vary considerably across individuals and reference identities.}
    \label{fig:jnd_ifs}
\end{figure}

Across all four identity groups including Indian male/female and Malaysian male/female (Figure~\ref{fig:jnd_ifs}), the 1-JND response distributions are consistently right-skewed, with the majority of responses concentrated within 0–5 frames of the reference image. However, the degree of spread varies notably by identity: Malaysian female and Malaysian male exhibit the sharpest peaks (response counts of $\sim 115$ and $\sim 90$, respectively), indicating tighter and more consistent perceptual thresholds, while Indian female and Indian male show broader distributions with heavier tails extending to 60–70 frames, reflecting greater inter-subject variability in perceptual sensitivity. These differences underscore that perceptual thresholds are identity-dependent and cannot be assumed uniform across individuals, motivating the need for subject-level PAQ calibration.

\subsection{Statistical Analysis in Response Time}
\label{app:response_time}

\begin{figure}
    \centering
    \includegraphics[width=\linewidth]{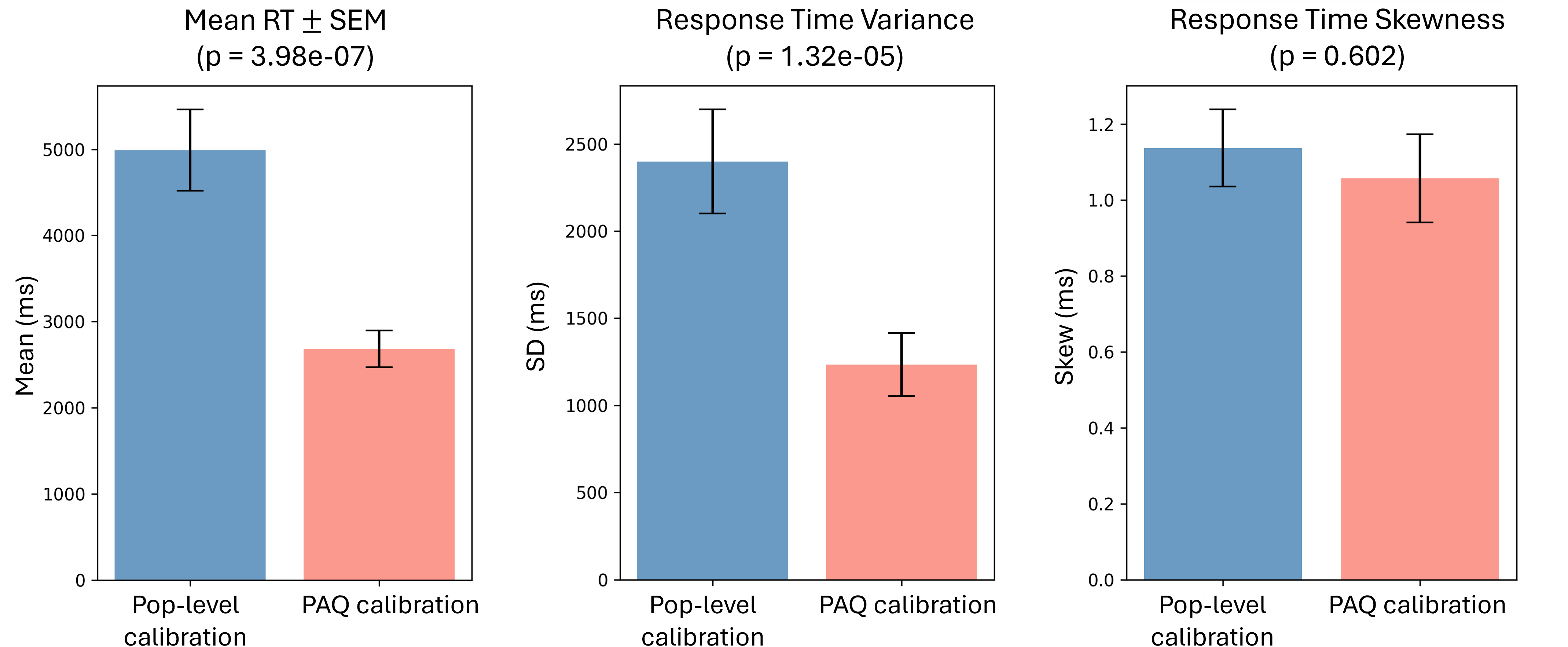}
    \vspace{-18pt}
    \caption{Group-level response time (RT) statistics comparing pop-level calibration and PAQ calibration conditions. PAQ calibration produced significant reductions in both mean RT and within-participant RT variability, while per-participant RT skewness remained statistically unchanged.}
    \label{fig:group__rt}
\end{figure}

\paragraph{Group-Level Descriptive Statistics} Group-level descriptive analyses examined whether PAQ calibration changed the
central tendency, spread, and distributional shape of individual participants'
RT profiles, using paired $t$-tests on per-participant summary statistics
computed separately for the population-level calibrated and PAQ calibrated conditions. As shown in Figure~\ref{fig:group__rt}, mean RT dropped
sharply and significantly from approximately 5.00\,s ($\pm$0.45\,SEM)
before calibration to 2.65\,s ($\pm$0.20\,SEM) after
($p = 3.98 \times 10^{-7}$), a reduction of roughly 47\%, with the SEM itself
nearly halving in the PAQ calibrated condition, indicating not only a lower group
average but also substantially greater consistency across participants in their
overall response speed. Intra-individual RT variability, indexed by
within-participant standard deviation, showed a parallel reduction of
comparable magnitude: group mean SD fell from ${\sim}2.38$\,s
($\pm$0.27\,SEM) to ${\sim}1.23$\,s ($\pm$0.20\,SEM),
$p = 1.32 \times 10^{-5}$ (${\approx}48\%$ decrease), confirming that PAQ
calibration suppressed trial-to-trial RT fluctuations within each participant
as well as overall level. Notably, the proportional reductions in mean and SD
are nearly identical (${\approx}47$--$48\%$), implying that the coefficient of
variation ($\mathrm{SD}/\mathrm{Mean}$) was approximately preserved across
conditions: calibration scaled the entire RT distribution downward without
fundamentally altering each participant's relative internal consistency. In
contrast, the distributional shape of individual RT profiles, quantified as
per-participant skewness, remained statistically unchanged between conditions
(Before PAQ: $1.13 \pm 0.09$; After PAQ: $1.06 \pm 0.08$; $p = 0.602$). Both
conditions exhibited moderate positive skew of approximately 1.1, meaning each
participant's RT distribution retained its characteristic right-tailed form
after calibration. This dissociation between the large reductions in mean and
SD on the one hand, and the stability of per-participant skewness on the other indicates that PAQ calibration rescales and
compresses individual RT distributions while preserving their shape. 

    \paragraph{Per-Position Wilcoxon Signed-Rank Tests Along the Affect Spectrum} To test whether the before-to-after PAQ RT reduction was statistically reliable at each individual position along the affect spectrum, we applied paired Wilcoxon signed-rank tests (one per position, $N = 52$ paired participants, one-sided: Before $>$ After) with Bonferroni correction for 16 simultaneous comparisons. The full results are reported in the table~\ref{tab:rt_wilcoxon_a} and table~\ref{tab:rt_wilcoxon_b}. All 16 positions reached significance after Bonferroni correction, confirming that the RT reduction following PAQ calibration is robust and pervasive across the entire affect spectrum rather than being driven by a subset of positions. The largest absolute reductions were concentrated at the emotionally extreme endpoints, $\text{happy}_{88}$ yielded the single greatest mean decrease (-3.76 s), followed by $\text{sad}_{66}$ (-2.928 s) and $\text{sad}_{88}$ (-2.729 s), consistent with the interpretation that population-level calibration most severely inflates RTs when stimuli fall outside a participant's natural affective reference range, an effect that is most pronounced for the highest-intensity morphs. The smallest reductions were observed at the near-neutral, mid-range positions, particularly $\text{happy}_{55}$ (-0.965 s) and $\text{sad}_{55}$/$\text{sad}_{44}$ ($\sim$-1.73-1.76 s), which also returned the weakest (though still corrected-significant) p values ($p_\text{bonf}\approx 0.005–0.014$). This gradient, largest calibration benefit at the emotionally clear extremes, smallest at the ambiguous midpoints, which suggests that perceptual uncertainty about mid-range morphs is less sensitive to individual reference-frame anchoring, because these stimuli are inherently difficult to categorize regardless of calibration.

\begin{table}[htbp]
\centering
\begin{tabular}{lrrr}
\toprule
Position & Mean $\Delta$ (s) & Median $\Delta$ (s) & W \\
\midrule
sad\_88   & -2.729 & -1.387 & 1.182 \\
sad\_77   & -2.173 & -1.325 & 1.210 \\
sad\_66   & -2.928 & -1.916 & 1.248 \\
sad\_55   & -1.731 & -0.684  & 1.060 \\
sad\_44   & -1.759 & -1.125 & 1.111 \\
sad\_33   & -2.232 & -1.319 & 1.132 \\
sad\_22   & -2.516 & -2.043 & 1.225 \\
sad\_11   & -2.101 & -0.999  & 1.160 \\
happy\_11 & -2.134 & -1.189 & 1.194 \\
happy\_22 & -2.531 & -1.246 & 1.084 \\
happy\_33 & -2.569 & -1607 & 1243 \\
happy\_44 & -2.235 & -1.339 & 1.162 \\
happy\_55 & -0.965  & -0.810  & 1.062 \\
happy\_66 & -2.563 & -1.613 & 1.253 \\
happy\_77 & -1.998 & -1.082 & 1.033 \\
happy\_88 & -3.760 & -1.845 & 1.241 \\
\bottomrule
\end{tabular}
\vspace{5pt}
\caption{Wilcoxon signed-rank test: reaction time differences ($\Delta$) across facial expression positions (descriptive statistics).}
\label{tab:rt_wilcoxon_a}
\end{table}

\begin{table}[htbp]
\centering
\begin{tabular}{lrrr}
\toprule
Position & $p$ (raw) & $p$ (Bonf.) & Sig. \\
\midrule
sad\_88   & $3.57 \times 10^{-6}$ & $5.71 \times 10^{-5}$ & $\checkmark$ \\
sad\_77   & $1.04 \times 10^{-6}$ & $1.67 \times 10^{-5}$ & $\checkmark$ \\
sad\_66   & $1.83 \times 10^{-7}$ & $2.92 \times 10^{-6}$ & $\checkmark$ \\
sad\_55   & $3.64 \times 10^{-4}$ & $5.83 \times 10^{-3}$ & $\checkmark$ \\
sad\_44   & $6.07 \times 10^{-5}$ & $9.72 \times 10^{-4}$ & $\checkmark$ \\
sad\_33   & $2.74 \times 10^{-5}$ & $4.38 \times 10^{-4}$ & $\checkmark$ \\
sad\_22   & $5.27 \times 10^{-7}$ & $8.43 \times 10^{-6}$ & $\checkmark$ \\
sad\_11   & $8.96 \times 10^{-6}$ & $1.43 \times 10^{-4}$ & $\checkmark$ \\
happy\_11 & $2.12 \times 10^{-6}$ & $3.40 \times 10^{-5}$ & $\checkmark$ \\
happy\_22 & $1.61 \times 10^{-4}$ & $2.57 \times 10^{-3}$ & $\checkmark$ \\
happy\_33 & $2.26 \times 10^{-7}$ & $3.62 \times 10^{-6}$ & $\checkmark$ \\
happy\_44 & $8.25 \times 10^{-6}$ & $1.32 \times 10^{-4}$ & $\checkmark$ \\
happy\_55 & $3.41 \times 10^{-4}$ & $5.45 \times 10^{-3}$ & $\checkmark$ \\
happy\_66 & $1.40 \times 10^{-7}$ & $2.24 \times 10^{-6}$ & $\checkmark$ \\
happy\_77 & $8.66 \times 10^{-4}$ & $1.39 \times 10^{-2}$ & $\checkmark$ \\
happy\_88 & $2.49 \times 10^{-7}$ & $3.99 \times 10^{-6}$ & $\checkmark$ \\
\bottomrule
\end{tabular}
\vspace{5pt}
\caption{Wilcoxon signed-rank test: $p$-values with Bonferroni correction across facial expression positions.}
\label{tab:rt_wilcoxon_b}
\end{table}

\section{Facial Generation Sequences Across Synthesis Methods}\label{app:face_gen}

We investigate four facial expression generation methods for producing dynamic face sequences for PAQ: AU-based, GAN-based, diffusion-based and GenAI prompting. This section describes the implementation of each method, demonstrates the generated facial sequences for PAQ, and analyzes the failures that motivate our final generation method: VLM prompting.

\subsection{AU-based facial generation}
\begin{figure}
    \centering
    \includegraphics[width=\linewidth]{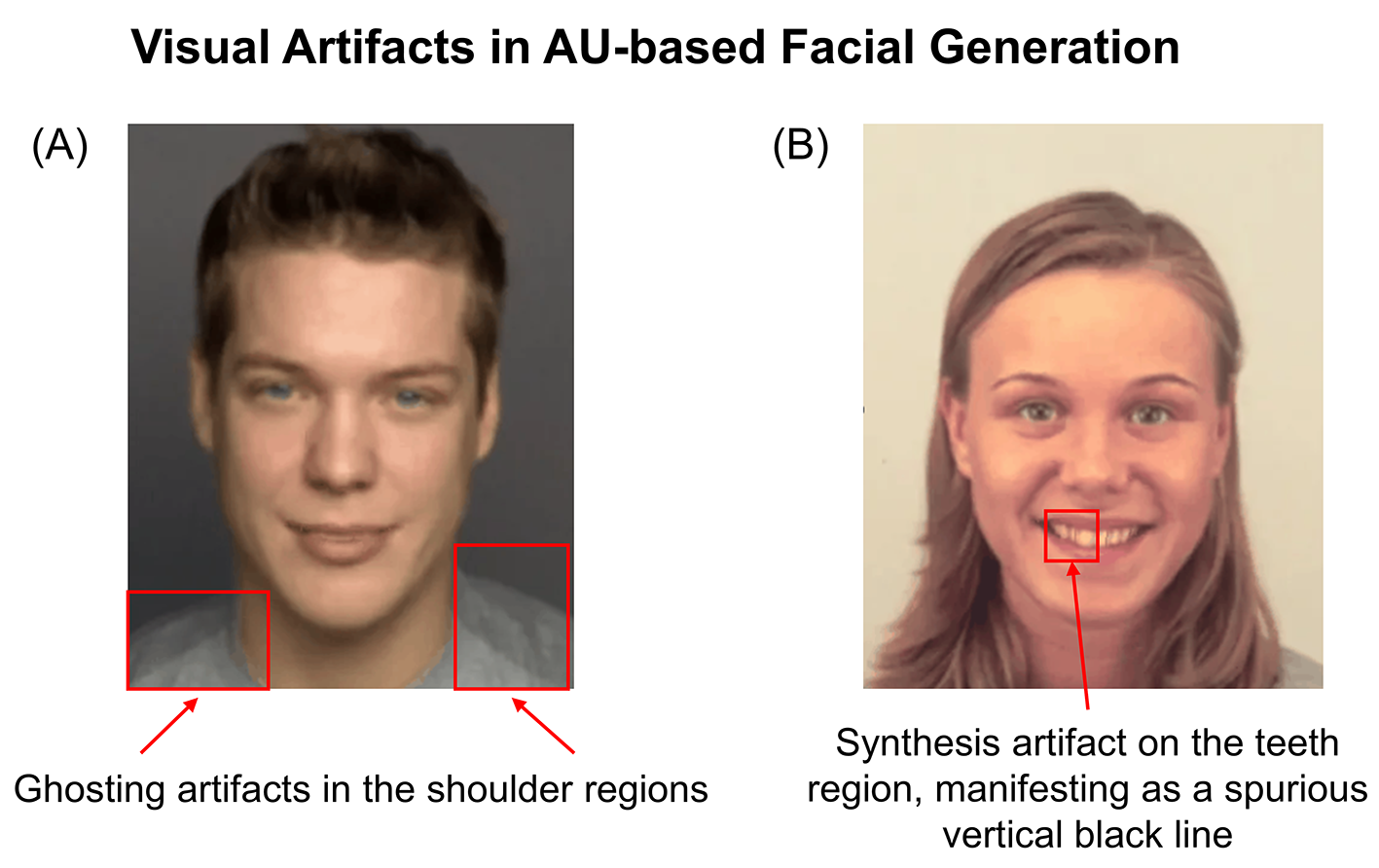}
    \vspace{-18pt}
    \caption{Visual artifacts in AU-based expression generation. (A) Ghosting artifacts in the bilateral shoulder regions,caused by the body positioning mismatch between source and target facial images. (B) Synthesis artifact in the teeth region, manifesting as a spurious vertical black line, indicative of the generator's failure to render fine-grained facial details faithfully under high-intensity AU activation. }
    \label{fig:au}
\end{figure}

The AU-based method performs geometry-driven morphing between two real faces. We proceed in five stages in facial morphing pipeline. First, facial landmarks are detected on both the source (neutral) and target (expressive: happy or sad) images using MediaPipe FaceMesh, augmented with boundary anchors along the image border to stabilize the outer region. Second, a Delaunay triangulation is computed once on the source landmarks and corresponding triangle indices are used across all intermediate frames to ensure topological consistency and to prevent fold-overs or distortions during warping. Third, for each morph ratio $\alpha \in [0,1]$, the intermediate landmark positions are computed via per-landmark linear interpolation: $p_i(\alpha) = (1 - \alpha)p_i^{\text{source}} + \alpha p_i^{\text{target}}$. Fourth, affine warping is applied to each corresponding triangle pair to produce a morphed face image at each $\alpha$. Fifth, a composite mask combining the convex hull of facia landmarks (face mark) and a segmentation-based hair mask (thresholded, dilated, and feathered) is used to blend the morphed face with a cross-dissolved background, suppressing flickering at region boundaries. \textbf{Observed issues:} Despite the structured geometric approach, the method produced several visible artifacts (Figure~\ref{fig:au}). Ghosting appeared consistently around the teeth, shoulder, hair boundary, and neck regions. The face position drifted slightly along the morphing sequence. These artifacts arise from the fundamental limitation of pixel-space blending: triangulated warping cannot model the appearance change accompanying structural transformations such as mouth opening.

\subsection{GAN-based facial generation}

\begin{figure}
    \centering
    \includegraphics[width=\linewidth]{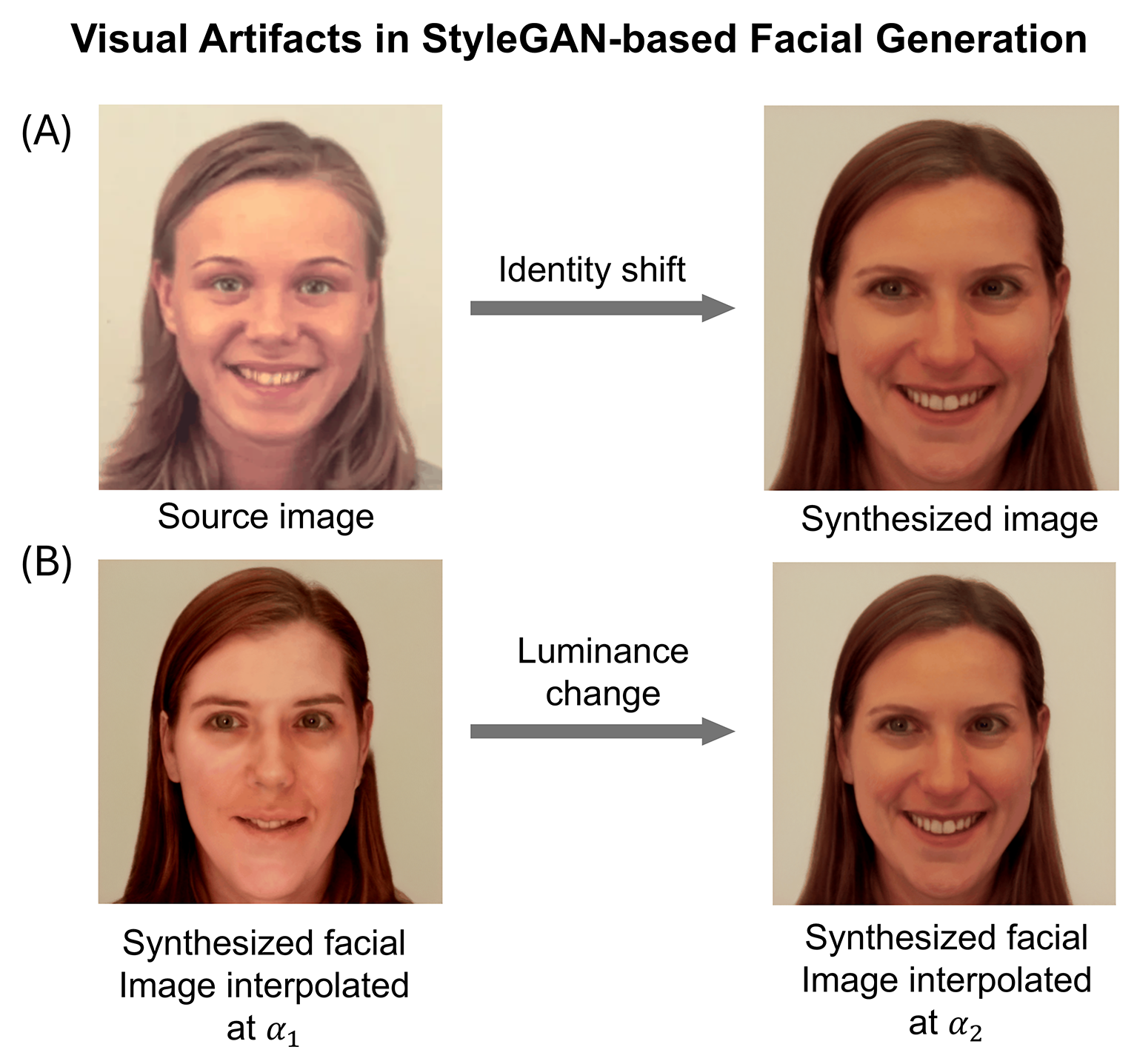}
    \vspace{-18pt}
    \caption{Visual artifacts in StyleGAN-based facial expression generation. (A) Identity shift: despite freezing the coarse $W+$ layers responsible for identity-related structure, the synthesized face (right) is perceptually different from the source image (left). (B) Luminance change: two synthesized frames at different $\alpha$ reveal noticeable brightness drift despite explicit Y-channel normalization. }
    \label{fig:styleGAN-art}
\end{figure}

\begin{figure}
    \centering
    \includegraphics[width=\linewidth]{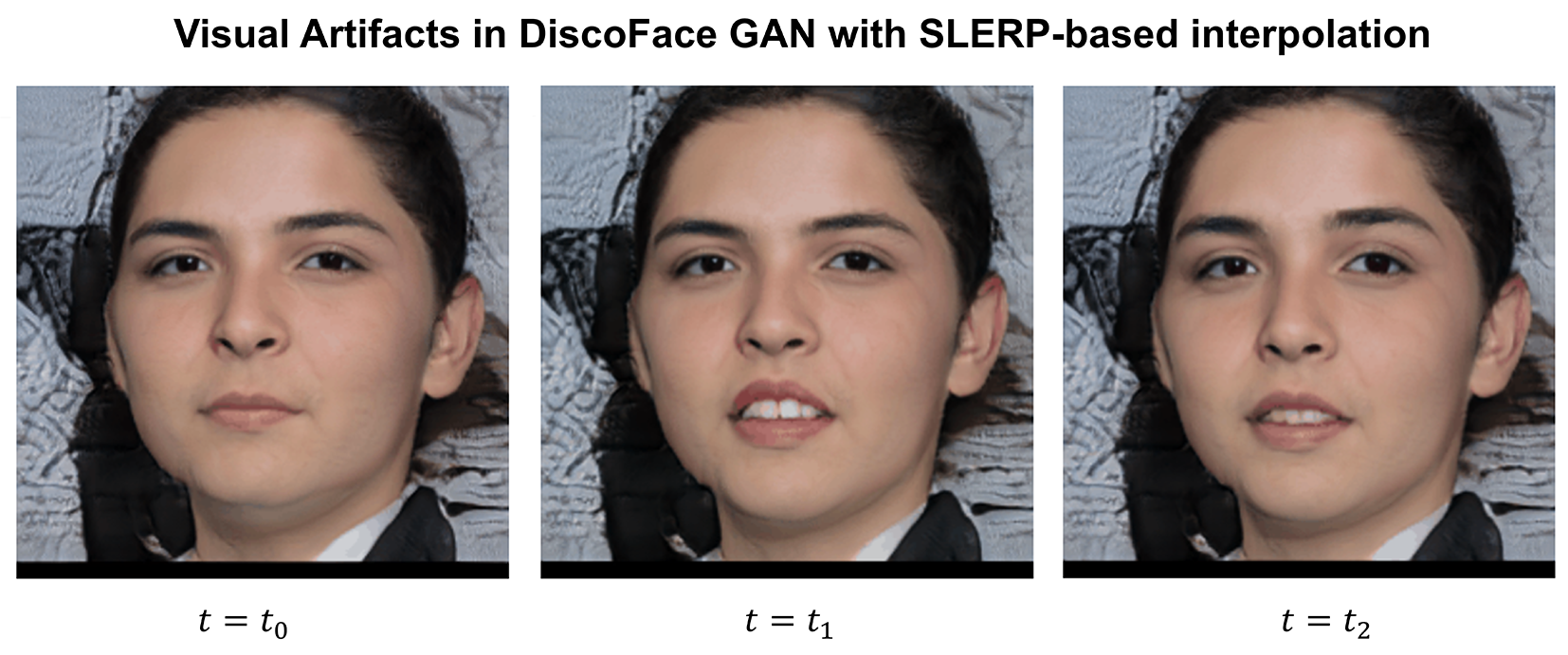}
    \vspace{-18pt}
    \caption{Example face sequence generated by DiscoFaceGAN with SLERP-based interpolation in the expression latent space. Despite equal step sizes in $t$, the perceived expression intensity changes non-uniformly across the three frames. The linear traversal in the EXP latent space is non-smooth, nonlinear, with small increments in $t$ yields perceptually negligible changes followed by abrupt, discontinuous shifs in expression.}
    \label{fig:dis_art}
\end{figure}

\paragraph{StyleGAN} StyleGAN introduces an intermediate lantent space $W$ between the input noise vector $Z$ and the synthesis network $G$. The mapping network $f: Z \rightarrow W$ produces a style vector $w$ that modulates each layer of the synthesis, enabling disentangled control over visual attributes such as hairstyle, pose, and expression at different spatial scales. To generate an expression sequence, each real face image is first inverted into the extended $W+$ space (a per-layer style code $w+ = (w_0, \ldots, w_{L-1}), w_i \in W$) using an encoder. The coarse layers, which control the identity-related structure, are frozen at the latent codes of the source images. Only the fine layers, which govern expression and texture details, are interpolated across $\alpha \in [0, 1]$ between the neutral and expressive latents. Each frame is then synthesized using deterministic noise to suppress temporal flickering, and luminance is normalized to a fixed mean/std in the Y channel to reduce brightness drift. \textbf{Observed issues:} Despite luminance normalization, the output frames exhibited noticeable identity and luminance changes (Figure~\ref{fig:styleGAN-art}). The pretrained StyleGAN2-ADA checkpoint was trained on a general face distribution and does not guarantee faithful reconstruction of any given individual's appearance, leading to systematic identity drift throughout the sequence.
\paragraph{DiscoFace GAN}
DiscoFaceGAN extends the StyleGAN framework with an explicit semantic latent structure designed for face attribute disentanglement. The input latent $z (179-dim)$ is decomposed into four semantic blocks: identity (ID; 128), expression (EXP:32), lightning (GEMMA: 16), and pose (ROT:3). A mapping network bridges this semantic space to a richer $\lambda$ space (254-dim), which is then fed to the StyleGAN synthesis network. To generate an expression sequence for a single subject, identity, lighting, and pose are held fixed while the EXP block is varied. The expressive (happy or sad) direction in EXP space is extracted by projecting both the neutral and expressive reference images into the latent space, isolating the EXP slide from each, and computing the normalized difference vector $d_{\text{exp}} = e_H - e_N / \| e_H - e_N\|$. Animation frames are produced by scaling this direction by a scalar $s$ linearly increasing from 0 to $\sim$7 over the desired number of frames. \textbf{Observed issues:} Linear traversal of the EXP latent produced sequences where most frames appeared perceptually static, followed by a sudden visible shift in expression (Figure~\ref{fig:dis_art}). This reflects the nonlinear mapping between latent distance and perceived expression intensity: the generator's latent geometry is curved and entangled, human perception of facial action units is nonlinear, and equal latent step sized produce small perceptual changes at first and abrupt changes later. 
\subsection{Diffusion model for facial generation}
The diffusion-based method operates in the noise latent space of a pretrained unconditional DDPM. The pipeline consists of three stages.

\begin{algorithm}
\caption{DDIM-Invert: Solve for $z_T$ such that $\text{DECODE}_{\eta=0}(z_T) \approx x_{\text{tar}}$}
\label{alg:ddim-invert}
\begin{algorithmic}[1]
\Require Target image $x_{\text{tar}} \in [-1,1]^{3 \times H \times W}$; UNet $\varepsilon_\theta$; cumulative alphas $\{\bar{\alpha}_t\}_{t=0}^{T-1}$; inversion DDIM indices $\{\tau_i^{\text{inv}}\}_{i=0}^{K_{\text{inv}}-1}$; iterations $I$; learning rate $\eta$; patience iters $n$.
\Ensure Inverted terminal latent $z_T$.
\State Initialize a latent $z_0 \sim \mathcal{N}(0, I)$
\State Initialize the loss and set the best latent $(L_{\min}, z_\star) \leftarrow (+\infty, z_0)$
\For{iter $= 1$ \textbf{to} $I$}
    \State $x_0 \leftarrow \text{DECODE}_{\eta=0}\!\left(z_0,\, \{\tau_i^{\text{inv}}\},\, \{\bar{\alpha}_t\}\right)$
    \State $\mathcal{L} \leftarrow \|x_0 - x_{\text{tar}}\|_2^2$
    \State \textbf{Adam step:} $z_0 \leftarrow z_0 - \eta\,\nabla_z \mathcal{L}$
    \State \textbf{(optional)} clamp $\text{RMS}(z_0) \leq 2$ \Comment{stay on typical noise shell}
    \If{$\mathcal{L} < L_{\min}$}
        \State $(L_{\min}, z_\star) \leftarrow (\mathcal{L}, z_0)$ \Comment{autosave best}
    \EndIf
    \If{no improvement for $\geq n$} \textbf{break} \EndIf
\EndFor
\State \Return $z_\star$
\end{algorithmic}
\end{algorithm}

\paragraph{Stage 1: DDIM inversion} Each input image (neutral and expressive) is deterministically encoded into a terminal noise latent $z_T$ via DDIM inversion (Algorithm~\ref{alg:ddim-invert}), which iteratively optimizes a latent $z_0$ such that $DECODE_{\eta = 0}(z_0) \approx x_{\text{target}}$, using gradient descent with Adam and early stopping. This yields $z_N$ (neutral latent) and $z_H$ (happy latent).

\begin{algorithm}
\caption{SLERP Path Between Endpoints $z_N$ and $z_H$}
\label{alg:slerp}
\begin{algorithmic}[1]
\Require Endpoint latents $z_N, z_H \in \mathbb{R}^{C \times H \times W}$; number of frames $F$; small $\epsilon > 0$.
\Ensure Latent path $\{z(t_i)\}_{i=0}^{F-1}$.
\State Flatten-and-normalize: $\hat{a} \leftarrow z_N / \|z_N\|$, \quad $\hat{b} \leftarrow z_H / \|z_H\|$
\State $\theta \leftarrow \arccos\!\left(\mathrm{clip}(\hat{a}^\top \hat{b},\; {-1{+}\epsilon},\; {1{-}\epsilon})\right)$
\For{$i = 0$ \textbf{to} $F-1$}
    \State $t_i \leftarrow i / (F-1)$
    \If{$|\sin\theta| < 10^{-6}$} \Comment{nearly colinear}
        \State $z(t_i) \leftarrow (1 - t_i)\, z_N + t_i\, z_H$ \Comment{LERP fallback}
    \Else
        \State $z(t_i) \leftarrow \dfrac{\sin((1-t_i)\theta)}{\sin\theta}\, z_N + \dfrac{\sin(t_i\,\theta)}{\sin\theta}\, z_H$
    \EndIf
\EndFor
\State \Return $\{z(t_i)\}_{i=0}^{F-1}$
\end{algorithmic}
\end{algorithm}

\paragraph{Stage 2: Slerp interpolation} The two terminal latents are interpolated along the geodesic of the noise shell using Spherical Linear Interpolation (Algorithm~\ref{alg:slerp}): $z(\gamma) = \text{slerp}(z_N, z_H; \gamma) = [\sin(1 - \gamma)\theta/\sin \theta] z_N + [\sin(\gamma \theta) / \sin \theta] z_H,$ where $\theta = \arccos (\rangle z_N, z_H \rangle) / (\| z_N\| \|z_H\|).$ A linear fallback (LERP)  is used when the two latents are nearly collinear.

\begin{algorithm}
\caption{Deterministic DDIM Decode $\text{DECODE}_{\eta=0}(z_T)$}
\label{alg:ddim-decode}
\begin{algorithmic}[1]
\Require Terminal latent $z_T$; UNet $\varepsilon_\theta$; cumulative alphas $\{\bar{\alpha}_t\}$; decode DDIM indices $\{\tau_i\}_{i=0}^{K-1}$.
\Ensure Predicted clean image $x_0$.
\State $x_{\tau_{K-1}} \leftarrow z_T$
\For{$i = K-1,\, K-2,\, \ldots,\, 0$}
    \State $t \leftarrow \tau_i$
    \State $\varepsilon \leftarrow \varepsilon_\theta(x_t, t)$
    \State $\hat{x}_0 \leftarrow \dfrac{x_t - \sqrt{1 - \bar{\alpha}_t}\,\varepsilon}{\sqrt{\bar{\alpha}_t}}$
    \If{$i > 0$} \Comment{$\eta{=}0$ update}
        \State $x_{\tau_{i-1}} \leftarrow \sqrt{\bar{\alpha}_{\tau_{i-1}}}\,\hat{x}_0 + \sqrt{1 - \bar{\alpha}_{\tau_{i-1}}}\,\varepsilon$
    \Else
        \State \Return $\hat{x}_0$
    \EndIf
\EndFor
\end{algorithmic}
\end{algorithm}

\paragraph{Stage 3: Deterministic decoding} Each interpolated latent $z(\gamma)$ is decoded back to pixel space via deterministic DDIM ($\eta = 0$), marching from $t = T$ to $t = 0$ (Algorithm~\ref{alg:ddim-decode}).

\begin{figure}
    \centering
    \includegraphics[width=\linewidth]{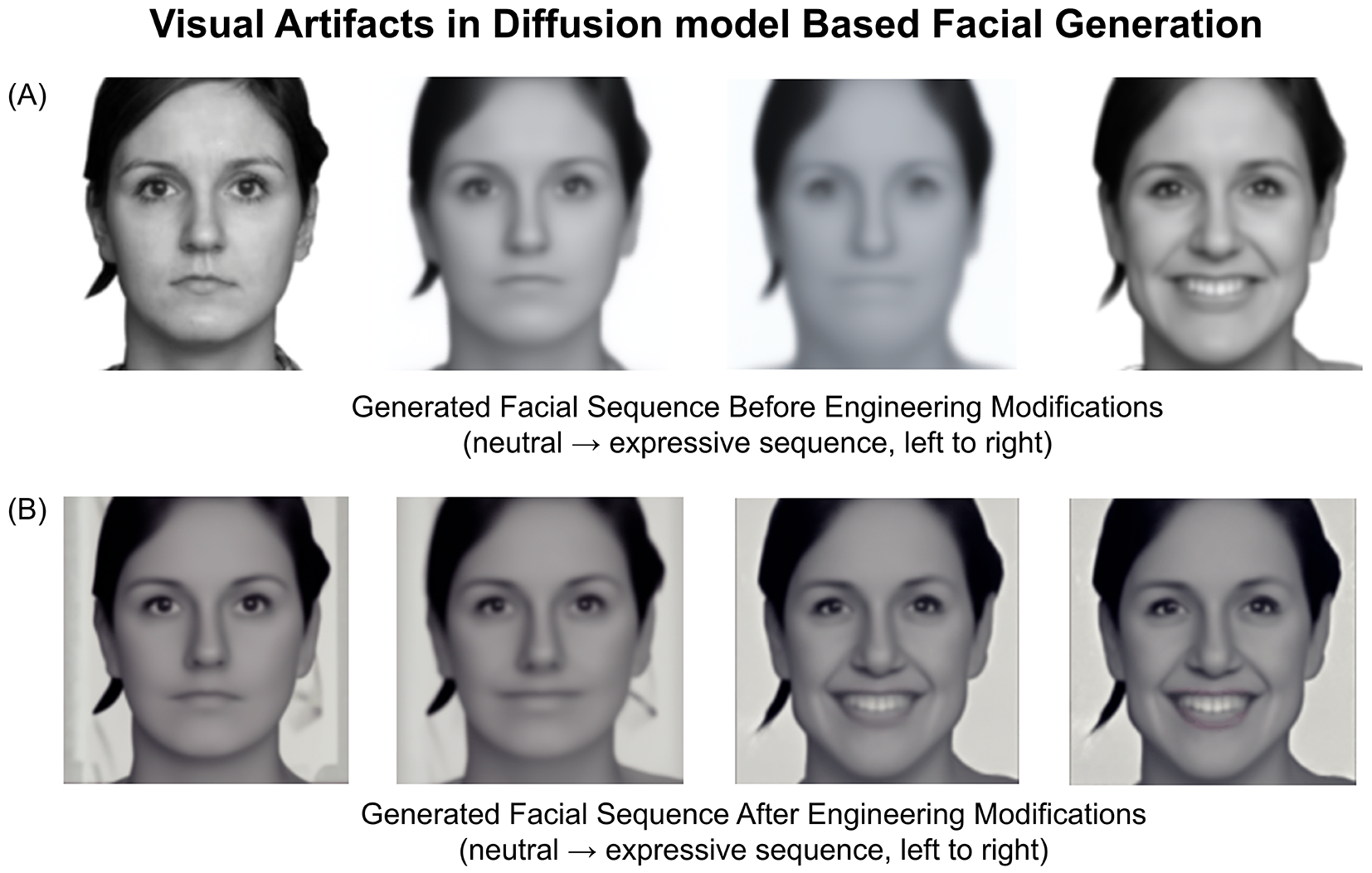}
    \vspace{-18pt}
    \caption{Visual artifacts in diffusion model-based expression generation, shwon as a sampled four-frame sequence interpolated along a SLERP arc from neutral (leftmost) to expressive (rightmost). (A) Before engineering modifications, intermediate frames exhibit severe blurriness and a characteristic "vague-then-clear" temporal profile. (B) After applying three engineering modifications, residual blurriness and artifacts as red lips around the mouth persist although local constant was improved.}
    \label{fig:ddim}
\end{figure}

Three engineering modifications were introduced to address the blurriness of intermediate frames: (1) a nonlinear DDIM step schedule with steps denser near $x_0$, so that $\sqrt{\tilde{\alpha}}$ changes linearly along the schedule, allocating more steps to the high frequency detail regime; (2) increased decode steps ($> 70$) for interpolated frames, preventing averaged textures at midpoints on the Slerp arc; and (3) dynamic thresholding on $\hat{x}_0$ predictions at $p = 0.995$, rescaling per-frame extremes to preserve local contrast and suppress the ``veil" artifact over fine details. \textbf{Observed issues:} Even with all three modifications, intermediate frames exhibited pronounced blurriness and a characteristic ``vague-then-clear" temporal profile, wherein decoded images became increasingly diffuse toward the midpoint of the Slerp are before sharpening again near the target (Figure~\ref{fig:ddim}). That is an intrinsic limitation of interpolating in the diffusion model's noise space: the model was not trained on paired neutral/expressive sequences, and the DDIM inversion does not guarantee that the reconstructed latent faithfully captures the full perceptual identity of the subject. Ghosting artifacts around the mouth region persisted across all variants.

\subsection{GenAI Image Generation (ChatGPT-Based)}
\begin{figure}
    \centering
    \includegraphics[width=\linewidth]{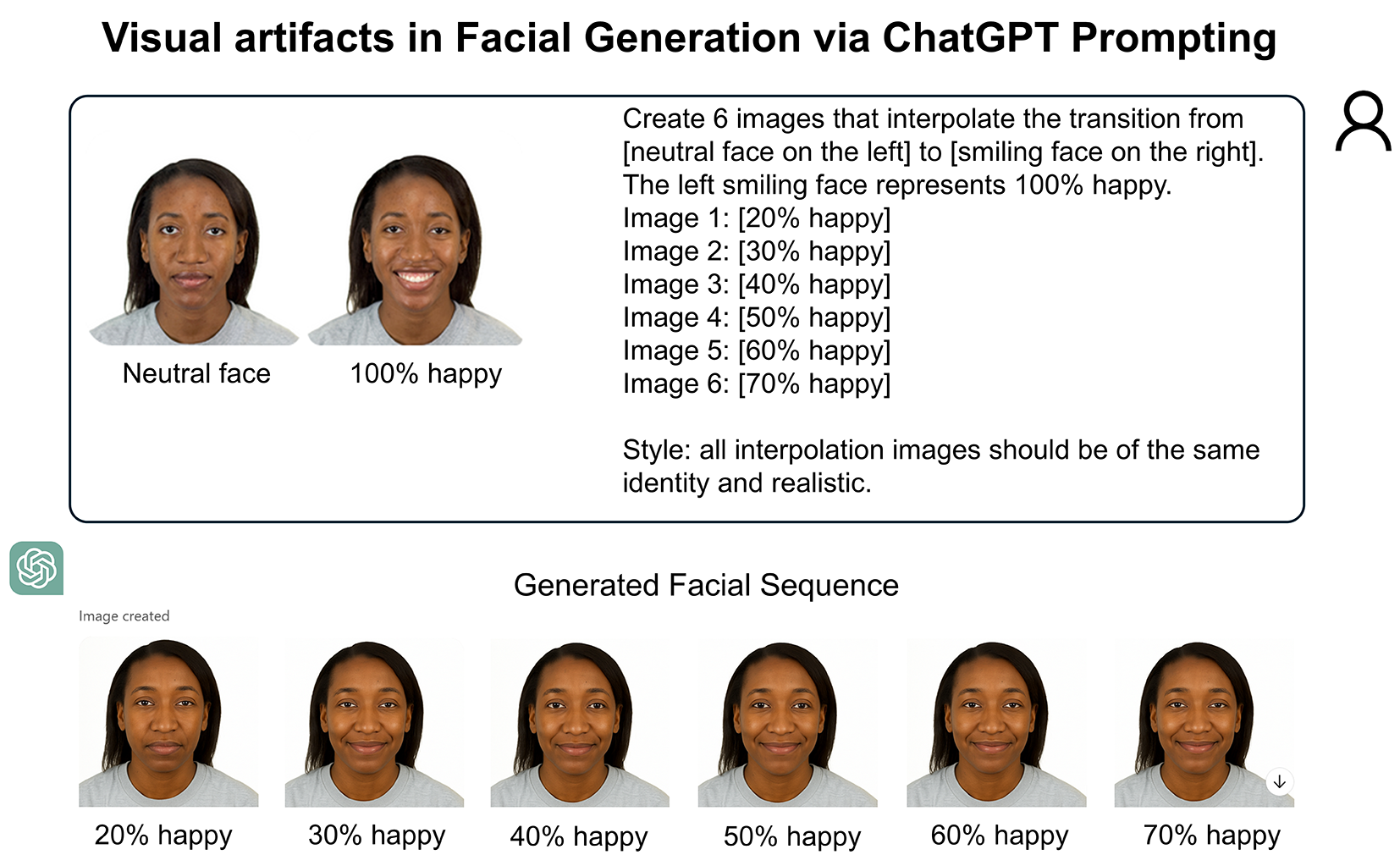}
    \vspace{-18pt}
    \caption{Facial expression generation via ChatGPT prompting. (\textbf{Top}) A structured prompt is provided with a neutral reference face and a 100\% happy face, requesting six interpolated frames ar explicitly labeled intensity levels (20\% - 70\% happy) with identity preservation. (\textbf{Bottom}) The six generated frames fail to produce perceptually linear or monotonically increasing expression intensity. Facial happiness does not scale proportionally with the specified percentages, and adjacent frames exhibit inconsistent perceptual jumps.}
    \label{fig:gpt_2}
\end{figure}

Prior to adopting video generation, we explored using multimodal large language models to directly synthesize intermediate expression frames. We input a structured prompt requested six images at explicitly labeled intensity levels (20\%, 30\%, \ldots, 70\% happy), with the instruction that all outputs should preserve the same identity and be realistic (Figure~\ref{fig:gpt_2}).
\textbf{Observed issues:} While the outputs appeared visually plausible at first glance, the shortcomings of this approach are threefold:
1. Identity changes across frames. The generated images consistently failed to preserve the subject's identity. Facial structure, skin tone, hair, and overall likeness drifted noticeably from the original reference, even within a single generation step. As shown above, the synthesized face is recognizably different from the input subject despite the explicit identity-preservation instruction. 2. Error compounds with chained generation. To build a fine-grained 100-frame slider, each generated frame must be used as the reference input for the next step. However, since each output already carries an identity shift, feeding it back into the model as a reference propagates and amplifies the error. Subsequent generations anchor to an already-drifted face, producing increasingly incorrect or low-quality results as the sequence progresses. 3. Inefficient and labor-intensive at scale. The resulting frame sequences are non-smooth, with visible perceptual jumps between adjacent frames. Constructing a single 100-frame slider requires 100 separate model calls, each of which must be individually inspected and curated by a human annotator. Scaled across 8 identities and 2 expressions each, this amounts to an impractical level of manual labor that cannot be systematically automated or made artifact-free.

\end{document}